%% file: main.tex
\pdfoutput=1
\documentclass[11pt,a4paper]{article}
\usepackage[hyperref]{acl2018}
\usepackage{times}
\usepackage{latexsym}

\usepackage[utf8]{inputenc}
\usepackage[]{amsmath}
\usepackage[]{amssymb}
\usepackage{array, rotating, multirow}
\usepackage{colortbl}

\usepackage{url}
\usepackage{pgf}

\aclfinalcopy % Uncomment this line for the final submission
\def\aclpaperid{422} %  Enter the acl Paper ID here

\title{Evaluating neural network explanation methods using \\ hybrid documents and morphosyntactic agreement}

\author{Nina Poerner, Benjamin Roth \& Hinrich Sch{\"u}tze \\
  Center for Information and Language Processing \\
  LMU Munich, Germany \\
  {\tt poerner@cis.lmu.de}}

\date{}

\usepackage[normalem]{ulem}

\newcounter{notecounter}
\newcommand{\enotesoff}{\long\gdef\enote##1##2{}}
\newcommand{\enoteson}{\long\gdef\enote##1##2{{
\stepcounter{notecounter}
{\large\bf
\hspace{1cm}\arabic{notecounter} $<<<$ ##1: ##2
$>>>$\hspace{1cm}}}}}
\enoteson
\enotesoff

\def\figref#1{Fig~\ref{fig:#1}}

\def\tabref#1{Table~\ref{tab:#1}}

\def\secref#1{\S\ref{sec:#1}}

\def\eqref#1{Eq~\ref{eqn:#1}}

\def\eqlabel#1{\label{eqn:#1}}

\newcommand{\gr}{\rowcolor{gray!25}}
\newcommand{\rot}[1]{\rotatebox[origin=c]{90}{#1}}
\begin{document}
\maketitle

\input{abstract}

\input{example-manual-snippet}

\input{introduction}
\input{evaluation-paradigms}
\input{evaluation-paradigms-hybrid}
\input{evaluation-paradigms-morphosyntactic}

\input{example-morphosyntactic-snippet}

\input{explanation-methods}
\input{explanation-methods-derivative}
\input{explanation-methods-lrp}
\input{explanation-methods-deeplift}
\input{explanation-methods-gamma}
\input{explanation-methods-omit}
\input{explanation-methods-lime}
\input{experiments}
\input{experiments-hybrid}
\input{experiments-morphosyntactic}

\input{example-hybrid-limsse}

\input{discussion}
\input{related-work}
\input{summary}
\input{explanation-methods-implementation}
\section{Acknowledgement}
We gratefully acknowledge funding for this work by the European Research
Council (ERC \#740516).

\bibliography{tmp}
\bibliographystyle{acl_natbib}

\clearpage
\section{Supplementary material}
\input{appendix}
\input{example-morphosyntactic1}
\input{example-morphosyntactic2}

\input{example-morphosyntactic3}

\input{example-manual-newsgroup-cnn}
\input{example-manual-newsgroup-gru}
\input{example-hybrid-newsgroup-qgru}
\input{example-hybrid-newsgroup-lstm}
\input{example-hybrid-newsgroup-qlstm}
\input{example-hybrid-yelp-gru}
\input{example-hybrid-yelp-lstm}

\clearpage

\end{document}

%% file: abstract.tex
\begin{abstract}
The behavior of deep neural networks (DNNs) is hard to
understand.
This makes it necessary to explore post hoc explanation methods.
We conduct
the first comprehensive evaluation of explanation methods
for NLP. To this end, we design two
novel evaluation paradigms that cover two important classes of NLP problems:
small context and large context problems.
Both paradigms require no manual
annotation and are therefore broadly applicable.
We also introduce LIMSSE, an explanation
method inspired by LIME that is designed for NLP. We show empirically
that LIMSSE, LRP and DeepLIFT are the most effective
explanation methods and recommend them for explaining DNNs
in NLP.
\end{abstract}

%% file: example-manual-snippet.tex
\begin{figure*}
\center
\scriptsize
\begin{tabular}{|m{0.05\textwidth}m{0.89\textwidth}|}
\hline 
${\mathrm{lrp}}$& \input{examples/12_CNN_lrp_short_138.tex} \\ \hline \hline
${{\mathrm{grad}^\mathrm{L2}_{1
p}}}$& \input{examples/QGRU_newsgroup_grad_prob_l2_short_630.tex} \\ \\[-7pt]
${{\mathrm{limsse}^\mathrm{ms}_s}}$& \input{examples/QGRU_newsgroup_lime_raw_short_630.tex} \\ \hline
\end{tabular}
\caption{\textbf{Top:} sci.electronics post (not hybrid). Underlined: Manual relevance ground truth. Green: evidence for sci.electronics.  Task method: CNN. \textbf{Bottom:} hybrid newsgroup post, classified talk.politics.mideast. Green: evidence for talk.politics.mideast. Underlined: talk.politics.mideast fragment. Task method: QGRU. Italics: OOV. Bold: $\mathrm{rmax}$ position. See supplementary for full texts.}
\label{fig:example-manual-snippet}
\end{figure*}

%% file: introduction.tex
\section{Introduction}
DNNs are complex models that combine linear transformations with different types of nonlinearities.
If the model is deep, i.e., has
many layers, then its behavior during training and
inference is notoriously hard to understand.

This is a problem for both scientific methodology and
real-world deployment.  Scientific methodology demands that
we understand our models.  In the real world,
a decision (e.g., ``your blog post is offensive and has
been removed'') by itself is
often insufficient; in addition, an explanation of the decision may be
required (e.g., ``our system flagged the following 
words as offensive'').  The European Union plans to
mandate that intelligent systems used for sensitive
applications provide such explanations
(European General Data Protection Regulation, expected 2018,
cf.\  \citet{goodman2016european}).

A number of post hoc
explanation methods for
DNNs have been proposed.  Due to the complexity
of the DNNs they explain, these methods are necessarily
approximations and come with their own sources of
error. At this point, it is not clear which of these methods
to use when reliable explanations for a specific DNN
architecture are needed.

\begin{table}[b]
  \footnotesize
\centering
  \begin{tabular}{l|l}
  tasks & sentiment analysis,\\
  &morphological prediction, \ldots\\\hline
task methods & CNN, GRU, LSTM, \ldots \\\hline
explanation methods & LIMSSE, LRP, DeepLIFT, \ldots\\\hline
evaluation paradigms & hybrid document,\\
&morphosyntactic agreement
\end{tabular}
\caption{Terminology with examples.}
\end{table}

\textbf{Definitions.} (i) A \textit{task method} 
solves an NLP problem, e.g., a GRU that predicts sentiment.

(ii) An \textit{explanation method} explains the behavior of
a task method on a specific input.  For our
purpose, it is a function $\phi(t, k, \mathbf{X})$ that
assigns real-valued relevance scores for a target class $k$
(e.g., positive)
to positions $t$ in an input text $\mathbf{X}$ (e.g.,
``great food''). For this example, an explanation
method might assign:
$\phi(1,
k, \mathbf{X}) > \phi(2, k, \mathbf{X})$.

(iii) An \textit{(explanation) evaluation paradigm}
quantitatively evaluates explanation methods for a task method, e.g., by assigning them accuracies.

\textbf{Contributions.} (i) We present novel evaluation
paradigms for explanation methods for two classes of common
NLP tasks (see \secref{evaluation-paradigms}).
Crucially, \emph{neither paradigm requires manual
annotations} and our methodology is therefore broadly applicable.

(ii) Using these paradigms, we perform a
comprehensive evaluation of explanation methods
for NLP (\secref{explanation-methods}).  We
cover the most important classes of task methods,
RNNs and CNNs, as well as the recently proposed Quasi-RNNs.

(iii) We introduce LIMSSE (\secref{explanation-methods-lime}), an explanation
method inspired by LIME \cite{ribeiro2016should} that is designed for word-order sensitive task methods (e.g., RNNs, CNNs).
We show empirically that
LIMSSE, LRP \cite{bach2015pixel} and DeepLIFT \cite{shrikumar2017learning} are the most effective explanation methods (\secref{experiments}): LRP and DeepLIFT are the most consistent methods, while LIMSSE wins the hybrid document experiment.

%% file: evaluation-paradigms.tex
\section{Evaluation paradigms}
\label{sec:evaluation-paradigms}
In this section, we introduce two
novel evaluation
paradigms for explanation methods on two types of common
NLP tasks, \emph{small context} tasks and \emph{large
  context} tasks.
Small context tasks are defined as those that can be solved
by finding short, self-contained indicators, such as words and phrases,
and weighing them up (i.e., tasks where CNNs with pooling can be expected to perform well).
We design the \emph{hybrid document paradigm} for
evaluating explanation methods on small context tasks.
Large context tasks require the correct handling of long-distance
dependencies, such as subject-verb agreement.\footnote{Consider deciding the number of \emph{[verb]} in ``the children in the green house said that the big telescope \emph{[verb]}'' vs. ``the children in the green house who broke the big telescope \emph{[verb]}''. The local contexts of ``children'' or ``\emph{[verb]}'' do not suffice to solve this problem, instead, the large context of the entire sentence has to be considered.}
We design the \emph{morphosyntactic agreement paradigm} for
evaluating explanation methods on large context tasks.

We could also use
\textbf{human judgments}
for evaluation.
While we use \citet{mohseni2018human}'s manual relevance benchmark for
comparison, there are two issues with it:
(i) Due to the \emph{cost of human labor}, it is limited in size
and domain. 
(ii) More importantly, \emph{a good explanation
method should not reflect what humans attend to, but what
task methods attend to.} 
For instance, the family name ``Kolstad'' has 11 out of its 13 appearances in
the 20 newsgroups corpus in sci.electronics posts.
Thus, task methods probably learn it as a sci.electronics indicator.
Indeed, the explanation method
in \figref{example-manual-snippet} (top)
marks ``Kolstad'' as relevant, but the human annotator does not.

%% file: evaluation-paradigms-hybrid.tex
\subsection{Small context: Hybrid document paradigm}
\label{sec:evaluation-paradigms-hybrid}
Given a collection of documents, hybrid documents are
created by randomly concatenating
document fragments. 
We assume that, on average, the most relevant input for a
class $k$ in a hybrid document is located in a fragment that
stems from a document with gold label $k$.  Hence, an
explanation method succeeds if it places maximal relevance
for $k$ inside the correct fragment.

Formally, let $x_t$ be a word inside hybrid document
$\mathbf{X}$ that originates from a document $\mathbf{X'}$
with gold label $y(\mathbf{X'})$.
$x_t$'s gold label $y(\mathbf{X}, t)$ is set to $y(\mathbf{X'})$.
Let $f(\mathbf{X})$ be the class assigned to the hybrid document by a task method, and let $\phi$ be an explanation method as defined above.
Let $\mathrm{rmax}(\mathbf{X}, \phi)$ denote the position of the maximally relevant word in $\mathbf{X}$ for the predicted class $f(\mathbf{X})$.
If this maximally relevant word comes from a document with the correct gold label, the explanation method is awarded a hit:
\begin{equation}
\eqlabel{hit}
\mathrm{hit}(\phi, \mathbf{X}) =
\mathbb{I}[{y\big(\mathbf{X}, {\mathrm{rmax}(\mathbf{X}, \phi)}\big) = f(\mathbf{X})}]
\end{equation}
where $\mathbb{I}[P]$ is 1 if $P$ is true and 0 otherwise.
In \figref{example-manual-snippet} (bottom), the explanation method $\mathrm{grad}^\mathrm{L2}_{1 p}$ places $\mathrm{rmax}$ outside the correct (underlined) fragment.
Therefore, it does not get a hit point, while $\mathrm{limsse}^\mathrm{ms}_s$ does.

The pointing game accuracy of an explanation method is calculated as its total number of hit points divided by the number of possible hit points.
This is a form of the pointing game paradigm from computer vision \cite{zhang2016top}.

%% file: evaluation-paradigms-morphosyntactic.tex
\subsection{Large context: Morphosyntactic agreement paradigm}
\label{sec:evaluation-paradigms-morphosyntactic}
Many natural languages display morphosyntactic agreement between
words $v$ and $w$.
A DNN that predicts the agreeing feature in $w$ should
pay attention to $v$.
For example, in the sentence 
``the children with the telescope are home'',
the number of the verb (plural for ``are'') can be predicted from
the subject (``children'') without looking at the verb.
If the language allows for $v$ and $w$ to be far apart 
(\figref{example-hybrid-limsse}, top), successful task methods
have to be able to handle large contexts.

\citet{linzen2016assessing}
show that English verb number can be predicted by a unidirectional LSTM
with accuracy $> 99\%$, based on left context alone.
When a task method predicts the
correct number, we expect successful explanation methods to
place maximal relevance on the subject:
\begin{equation}
\mathrm{hit}_{\mathrm{target}}(\phi, \mathbf{X}) = 
\mathbb{I}[\mathrm{rmax}(\mathbf{X}, \phi) = \mathrm{target}(\mathbf{X})]
\nonumber
\end{equation}
where $\mathrm{target}(\mathbf{X})$ is the location of the subject, and $\mathrm{rmax}$ is calculated as above.
Regardless of whether the prediction is correct, we expect $\mathrm{rmax}$ to fall onto a noun that has the predicted number:
\begin{equation}
\mathrm{hit}_{\mathrm{feat}}(\phi, \mathbf{X}) = 
\mathbb{I}[\mathrm{feat}\big(\mathbf{X}, \mathrm{rmax}(\mathbf{X}, \phi)\big) = f(\mathbf{X})]
\nonumber
\end{equation}
where $\mathrm{feat}(\mathbf{X},t)$ is the morphological feature (here: number) of $x_t$.
In \figref{example-morphosyntactic-snippet}, $\mathrm{rmax}$ on ``link'' gives a $\mathrm{hit}_\mathrm{target}$ point (and a $\mathrm{hit}_\mathrm{feat}$ point), $\mathrm{rmax}$ on ``editor'' gives a $\mathrm{hit}_\mathrm{feat}$ point.
${\mathrm{grad}^\mathrm{L2}_{\int s}}$ does not get any points as ``history'' is not a plural noun.

Labels for this task
can be automatically generated using part-of-speech 
taggers and parsers, which are available for many
languages.

%% file: example-morphosyntactic-snippet.tex
\begin{figure}
\centering
\scriptsize
\begin{tabular}{|ll|}
\hline
${\mathrm{grad}^\mathrm{dot}_{\int s}}$&\input{examples/GRU_salience_raw_integrated_2446} [encourages ...]\\ 
${\mathrm{lrp}}$&\input{examples/GRU_lrp_2446} [encourages ...]\\
${\mathrm{limsse}^\mathrm{bb}}$&\input{examples/GRU_lime_class_2446} [encourages ...]\\
\hline \hline
${\mathrm{grad}^\mathrm{L2}_{\int s}}$&\input{examples/LSTM_l2_raw_integrated_2109} [are ...] \\
${\mathrm{occ}_1}$&\input{examples/LSTM_occlusion_2109} [are ...] \\
${\mathrm{limsse}^\mathrm{ms}_s}$&\input{examples/LSTM_lime_raw_2109} [are ...]\\ \hline
\end{tabular}
\caption{\textbf{Top:} verb context classified
  {singular}. Green: evidence for {singular}.
  Task method: GRU. \textbf{Bottom:} verb context classified {plural}. Green: evidence for {plural}. Task method: LSTM. Underlined: subject. Bold: $\mathrm{rmax}$ position.}
\label{fig:example-morphosyntactic-snippet}
\end{figure}

%% file: explanation-methods.tex
\section{Explanation methods}
\label{sec:explanation-methods}
In this section, we define the explanation methods 
that will be evaluated.
For our purpose, explanation methods produce
word relevance scores $\phi(t,k,\mathbf{X})$,
which are specific to a given class $k$ and a given input
$\mathbf{X}$.  $\phi(t,k,\mathbf{X}) >
\phi(t',k,\mathbf{X})$ means that $x_t$ contributed
more than $x_{t'}$ to
the task method's (potential) decision to classify $\mathbf{X}$
as $k$.

%% file: explanation-methods-derivative.tex
\subsection{Gradient-based explanation methods}
\label{sec:explanation-methods-derivative}
Gradient-based explanation methods approximate the contribution of some DNN input $i$ to some output $o$ with $o$'s gradient with respect to $i$ \cite{simonyan2013deep}.
In the following, we consider two output functions $o(k, \mathbf{X})$, the unnormalized class score $s(k, \mathbf{X})$ and the class probability $p(k|\mathbf{X})$:
\begin{equation}
\displaystyle
	\eqlabel{softmaxs}
	s(k, \mathbf{X}) = \vec{w}_k \cdot \vec{h}(\mathbf{X}) + b_k
	\end{equation}
	\begin{equation}
	\displaystyle
	\eqlabel{softmaxp}
	p(k|\mathbf{X}) = \frac{\mathrm{exp}\big(s(k, \mathbf{X})\big)}{\sum_{k'=1}^K \mathrm{exp}\big(s(k', \mathbf{X})\big)}
\end{equation}
where $k$ is the target class,
$\vec{h}(\mathbf{X})$ the document representation (e.g.,
an RNN's final hidden layer), $\vec{w}_k$ (resp.\ $b_k$) $k$'s weight vector (resp.\ bias).
  
The simple gradient of $o(k, \mathbf{X})$ w.r.t. $i$ is:
\begin{equation}
\displaystyle
\mathrm{grad}_1(i, k, \mathbf{X}) = \frac{\partial o(k, \mathbf{X})}{\partial i}\eqlabel{partial}
\end{equation}
$\mathrm{grad}_1$ underestimates the importance of inputs
that saturate
a nonlinearity
\cite{shrikumar2017learning}.
To address this, \citet{sundararajan2017axiomatic} integrate over all gradients on a linear
interpolation $\alpha \in [0,1]$ between a baseline input $\bar{\mathbf{X}}$ (here: all-zero embeddings) and $\textbf{X}$:
\begin{eqnarray}
&\mathrm{grad}_{\int}(i, k, \mathbf{X})
	= \int_{\alpha=0}^1  \frac{\partial o(k,
          \mathbf{\bar{X}} + \alpha (\mathbf{X} -
  \mathbf{\bar{X}}))}{\partial i} \partial
        \alpha \nonumber \\
	& \approx \frac{1}{M} \sum_{m=1}^M \frac{\partial o(k, \mathbf{\bar{X}} + \frac{m}{M} (\mathbf{X} - \mathbf{\bar{X}}))}{\partial i}\eqlabel{int}
\end{eqnarray}
where $M$ is a big enough constant (here: 50).

In NLP, symbolic inputs (e.g., words) are often
represented as one-hot vectors
$\vec{x}_t \in \{1,0\}^{|V|}$ and 
embedded
via a real-valued matrix: $\vec{e}_t
= \mathbf{M} \vec{x}_t$.  Gradients are computed with
respect to individual entries of $\mathbf{E} =
[\vec{e}_1 \ldots \vec{e}_{|\mathbf{X}|}]$.
\citet{bansal2016ask}
and \citet{hechtlinger2016interpretation}
use the L2 norm
to
reduce vectors of gradients to single values:
\begin{equation}
\eqlabel{L2}
\mbox{$\phi_{\mathrm{grad}^\mathrm{L2}}(t, k, \mathbf{X}) = ||\mathrm{grad}(\vec{e}_t, k, \mathbf{E})||$}
\end{equation}
where $\mathrm{grad}(\vec{e}_t, k, \mathbf{E})$ is a vector of elementwise gradients w.r.t. $\vec{e}_t$.
\citet{denil2014extraction} use the dot product of the gradient vector and the embedding\footnote{For $\mathrm{grad}^\mathrm{dot}_{\int}$, replace $\vec{e}_t$ with $\vec{e}_t - \vec{\bar{e}}_t$. Since our baseline embeddings are all-zeros, this is equivalent.}, i.e., the gradient of the ``hot'' entry in $\vec{x}_t$: 
\begin{equation}
\displaystyle
\mbox{$\phi_{\mathrm{grad}^\mathrm{dot}}(t, k, \mathbf{X}) = \vec{e}_t \cdot \mathrm{grad}(\vec{e}_t, k, \mathbf{E})$}	
	\eqlabel{dot}
\end{equation}

We use  ``$\mathrm{grad}_1$'' for
\eqref{partial},
``$\mathrm{grad}_{\int}$'' for \eqref{int},
``$\mbox{}_p$'' for \eqref{softmaxp},
``$\mbox{}_s$'' for \eqref{softmaxs},
``L2'' for \eqref{L2} and ``dot'' for \eqref{dot}. This
gives us eight explanation methods:
${\mathrm{grad}_{1s}^\mathrm{L2}}$,
${\mathrm{grad}_{1p}^\mathrm{L2}}$,
${\mathrm{grad}_{1s}^\mathrm{dot}}$,
${\mathrm{grad}_{1p}^\mathrm{dot}}$,
${\mathrm{grad}_{\int s}^\mathrm{L2}}$,
${\mathrm{grad}_{\int p}^\mathrm{L2}}$,
${\mathrm{grad}_{\int s}^\mathrm{dot}}$,
${\mathrm{grad}_{\int p}^\mathrm{dot}}$.

%% file: explanation-methods-lrp.tex
\subsection{Layer-wise relevance propagation}
\label{sec:explanation-methods-lrp}
Layer-wise relevance propagation (LRP) is a
backpropagation-based explanation method developed for fully
connected neural networks and CNNs \cite{bach2015pixel} and later extended to LSTMs \cite{arras2017explaining}.
In this paper, we use Epsilon LRP (Eq 58, \citet{bach2015pixel}).
Remember that the activation of neuron $j$, $a_j$, is the sum of weighted upstream activations, $\sum_i a_i w_{i,j}$, plus bias $b_j$, squeezed through some nonlinearity.
We denote the pre-nonlinearity activation of $j$ as ${a'}_j$.
The relevance of $j$, $R(j)$, is distributed to upstream neurons $i$ proportionally to the
contribution that $i$ makes to ${a'}_j$ in the forward pass:
\begin{equation}
\eqlabel{lrp}
\displaystyle
	R(i) = \sum_j R(j) \frac{a_i w_{i,j}}{{a'}_j +\mathrm{esign}({a'}_j)}
\end{equation}
This ensures that relevance is conserved between layers, with the exception of relevance attributed to $b_j$.
To prevent numerical instabilities, $\mathrm{esign}(a')$ returns $-\epsilon$ if $a' < 0$ and $\epsilon$ otherwise. We set $\epsilon= .001$.
The full algorithm is:
\begin{equation}
\begin{aligned}
&R(L_{k'}) =
	s(k, \mathbf{X}) \mathbb{I}[{k'=k}] \\
&\text{... recursive application of \eqref{lrp} ...}\\
&\phi_{\mathrm{lrp}}(t, k, \mathbf{X}) = \sum_{j=1}^{\mathrm{dim}(\vec{e}_t)} R(e_{t,j}) \\
\end{aligned}
	\nonumber
\end{equation}
where $L$ is the final layer, $k$ the target class and $R(e_{t,j})$  the relevance of dimension $j$ in the $t$'th embedding vector.
For $\epsilon \rightarrow 0$ and provided that all nonlinearities up to the unnormalized class score are $\mathrm{relu}$, Epsilon LRP is equivalent to the product of input and raw score gradient (here: $\mathrm{grad}^\mathrm{dot}_{1 s}$) \cite{kindermans2016investigating}.
In our experiments, the second requirement holds only for CNNs.

Experiments by \citet{ancona2017unified}
(see \secref{related-work}) suggest that LRP does not work well for LSTMs if
all neurons -- including gates -- participate in backpropagation.
We therefore use \citet{arras2017explaining}'s modification and treat sigmoid-activated gates as time step-specific weights rather than neurons.
For instance, the relevance of LSTM candidate vector $\vec{g}_t$ is calculated from memory vector $\vec{c}_t$ and input gate vector $\vec{i}_t$ as

\begin{equation}
R(g_{t,d}) = R(c_{t,d})  \frac{g_{t,d} \cdot i_{t,d} }{c_{t,d} + \mathrm{esign}(c_{t,d})} \nonumber
\end{equation}
This is equivalent to applying \eqref{lrp} while treating $\vec{i}_t$ as a diagonal weight matrix.
The gate neurons in $\vec{i}_t$ do not receive any relevance themselves.
See supplementary material for formal definitions of Epsilon LRP for different architectures.

%% file: explanation-methods-deeplift.tex
\subsection{DeepLIFT}
\label{sec:explanation-methods-deeplift}
DeepLIFT \cite{shrikumar2017learning} is another backpropagation-based explanation method.
Unlike LRP, it does not explain $s(k, \mathbf{X})$, but $s(k, \mathbf{X})- s(k, \bar{\mathbf{X}})$, where $\bar{\mathbf{X}}$ is some  baseline input (here: all-zero embeddings).
Following \citet{ancona2018towards} (Eq 4), we use this backpropagation rule:
\begin{equation}
\displaystyle
R(i) = \sum_j R(j) \frac{a_i w_{i,j} - \bar{a}_i w_{i,j}}{a'_j - \bar{a}'_j + \mathrm{esign}(a'_j - \bar{a}'_j)}\nonumber
\end{equation}
where $\bar{a}$ refers to the forward pass of the baseline.
Note that the original method has a different mechanism for avoiding small denominators; we use $\mathrm{esign}$ for compatibility with LRP.
The DeepLIFT algorithm is started with $R(L_{k'}) = \big(s(k, \mathbf{X}) - s(k, \bar{\mathbf{X}})\big) \mathbb{I}[k' = k]$.
On gated (Q)RNNs, we proceed analogous to LRP and treat gates as weights.

%% file: explanation-methods-gamma.tex
\subsection{Cell decomposition for gated RNNs}
\label{sec:explanation-methods-gamma}
The cell decomposition explanation method for LSTMs \cite{murdoch2017automatic} decomposes the unnormalized class score $s(k, \mathbf{X})$ (\eqref{softmaxs}) into additive contributions.
For every time step $t$, we compute how much of $\vec{c}_t$ ``survives'' until the final step $T$ and contributes to $s(k,\mathbf{X})$.
This is achieved by applying all future forget gates $\vec{f}$, the final $\tanh$ nonlinearity, the final output gate $\vec{o}_T$, as well as the class weights of $k$ to $\vec{c}_t$.
We call this quantity ``net load of $t$ for class $k$'':
\begin{equation}
\displaystyle
\mathrm{nl}(t, k, \mathbf{X}) = \vec{w}_k \cdot \Big(\vec{o}_T \odot \mathrm{tanh}\big((\prod_{j={t+1}}^T \vec{f}_j) \odot \vec{c}_t \big)\Big)\nonumber
\end{equation}
where $\odot$ and $\prod$ are applied elementwise.
The relevance of $t$ is its gain in net load relative to $t-1$: $\phi_\mathrm{decomp}(t,k,\mathbf{X}) = \mathrm{nl}(t, k, \mathbf{X}) - \mathrm{nl}(t-1, k, \mathbf{X})$. 
For GRU, we change the definition of net load:
\begin{equation}
\displaystyle
\mathrm{nl}(t, k, \mathbf{X}) = \vec{w}_k \cdot \big((\prod_{j=t+1}^T \vec{z}_j) \odot \vec{h}_t\big) \nonumber
\end{equation}
where $\vec{z}$ are GRU update gates.

%% file: explanation-methods-omit.tex
\subsection{Input perturbation methods}
\label{sec:explanation-methods-omit}
Input perturbation methods assume that the removal or masking of relevant inputs changes the output \cite{zeiler2014visualizing}.
Omission-based methods remove inputs completely \cite{kadar2017representation}, while occlusion-based methods replace them with a baseline \cite{li2016understanding}.
In computer vision, perturbations are usually applied to patches, as neighboring pixels tend to correlate \cite{zintgraf2017visualizing}.
To calculate the $\mathrm{omit}_N$ (resp. $\mathrm{occ}_N$) relevance of word $x_t$, we delete (resp. occlude), one at a
time, all $N$-grams that contain $x_t$, and
average the change in the unnormalized class score from \eqref{softmaxs}:
\begin{eqnarray}
\nonumber	
&\phi_{[\mathrm{omit}|\mathrm{occ}]_N}(t,k,\mathbf{X}) = \sum_{j=1}^{N} \big[ s(k, [\vec{e}_1 \ldots \vec{e}_{|\mathbf{X}|}]) \\ \nonumber
&- s(k, [\vec{e}_1 \ldots \vec{e}_{t-N-1+j}] \Vert \mathbf{\bar{E}} \Vert [\vec{e}_{t+j} \ldots \vec{e}_{|\mathbf{X}|}]) \big] \frac{1}{N} \\ \nonumber
\end{eqnarray}
where $\vec{e}_t$ are embedding vectors, $\Vert$ denotes concatenation and $\mathbf{\bar{E}}$ is either a sequence of length zero ($\phi_\mathrm{omit}$) or a sequence of $N$ baseline (here: all-zero) embedding vectors ($\phi_\mathrm{occ}$).

%% file: explanation-methods-lime.tex
\subsection{LIMSSE: LIME for NLP}
\label{sec:explanation-methods-lime}
Local Interpretable Model-agnostic Explanations (LIME) \cite{ribeiro2016should} is a framework for explaining predictions of complex classifiers.
LIME approximates the behavior of classifier $f$ in the neighborhood of input $\mathbf{X}$ with an interpretable (here: linear) model.
The interpretable model is trained on samples $\mathbf{Z}_1 \ldots \mathbf{Z}_N$ (here: $N=3000$), which are randomly drawn from $\mathbf{X}$, with ``gold labels'' $f(\mathbf{Z}_1) \ldots f(\mathbf{Z}_N)$.

Since RNNs and CNNs respect word order, 
we cannot use the bag of words sampling method from the original description of LIME.
Instead, we introduce Local Interpretable Model-agnostic
Substring-based Explanations (LIMSSE).
LIMSSE uniformly samples a length $l_n$ (here: $1 \leq l_n \leq 6$) 
and a starting point $s_n$, which define the substring $\mathbf{Z}_n = [\vec{x}_{s_n} \ldots \vec{x}_{s_n+l_n-1}]$.
To the linear model, $\mathbf{Z}_n$ is represented by a binary vector $\vec{z}_n \in \{0,1\}^{|\mathbf{X}|}$, where $z_{n,t} = \mathbb{I}[s_n \leq t < s_n+l_n]$.

We learn a linear weight vector $\hat{\vec{v}}_k \in
\mathbb{R}^{|\mathbf{X}|}$, whose entries are word relevances for $k$, i.e., $\phi_{\mathrm{limsse}}(t,k,\mathbf{X}) = \hat{v}_{k,t}$.
To optimize it, we experiment with three loss functions.
The first, which we will refer to as $\mathrm{limsse}^\mathrm{bb}$, assumes that our DNN is a total black box that delivers
only a classification:
\begin{equation} \nonumber
	\begin{aligned}
		\hat{\vec{v}}_k & = \underset{\vec{v}_k}{\mathrm{argmin}} \sum_n - \big[\mathrm{log}\big(\sigma(\vec{z}_n \cdot \vec{v}_k)\big)\mathbb{I}[f(\mathbf{Z}_n) = k] \\
							    & + \mathrm{log}\big(1-\sigma(\vec{z}_n \cdot \vec{v}_k)\big) \mathbb{I}[f(\mathbf{Z}_n) \neq k]\big] \nonumber
	\end{aligned}
	\end{equation}
where $f(\mathbf{Z}_n) = \mathrm{argmax}_{k'} \big(p(k'|\mathbf{Z}_n)\big)$.
The black box approach is maximally general, but insensitive to the magnitude of evidence found in $\mathbf{Z}_n$.
Hence, we also test magnitude-sensitive loss functions:
\begin{equation}
\displaystyle
\hat{\vec{v}}_k = \underset{\vec{v}_k}{\mathrm{argmin}} \sum_n \big(\vec{z}_n \cdot \vec{v}_k - o(k, \mathbf{Z}_n)\big)^2\nonumber\\ \nonumber
\end{equation}
where $o(k, \mathbf{Z}_n)$ is one of $s(k, \mathbf{Z}_n)$ or $p(k|\mathbf{Z}_n)$.
We refer to these as $\mathrm{limsse}^\mathrm{ms}_s$ and $\mathrm{limsse}^\mathrm{ms}_p$.

%% file: experiments.tex
\newcolumntype{?}{!{\vrule width 2pt}}

\newcommand{\bs}[1]{\underline{\textbf{#1}}}
\newcommand{\gd}[1]{\underline{#1}}

\setlength\tabcolsep{1.9pt}
\begin{table*}
\centering
{\scriptsize
\begin{tabular}{l?ccccc|ccccc?ccccc?cccc|cccc|cccc}
column&\tiny C01&\tiny C02&\tiny C03&\tiny C04&\tiny C05&\tiny C06&\tiny C07&\tiny C08&\tiny C09&\tiny C10&\tiny C11&\tiny C12&\tiny C13&\tiny C14&\tiny C15&\tiny C16&\tiny C17&\tiny C18&\tiny C19&\tiny C20&\tiny C21&\tiny C22&\tiny C23&\tiny C24&\tiny C25&\tiny C26&\tiny C27\\\hline
&\multicolumn{10}{c?}{hybrid document experiment}&\multicolumn{5}{c?}{man. groundtruth}&\multicolumn{12}{c}{morphosyntactic agreement experiment} \\ \cline{2-28}
&&&&&&&&&&&&&&&&\multicolumn{4}{c|}{$\mathrm{hit}_\mathrm{target}$}&\multicolumn{8}{c}{$\mathrm{hit}_\mathrm{feat}$} \\
&\multicolumn{5}{c|}{yelp}&\multicolumn{5}{c?}{20 newsgroups}&\multicolumn{5}{c?}{20 newsgroups}&\multicolumn{8}{c|}{$f(\mathbf{X}) = y(\mathbf{X})$}&\multicolumn{4}{c}{$f(\mathbf{X}) \neq y(\mathbf{X})$} \\ \cline{2-28}
$\phi$&\rotatebox{90}{\tiny
GRU}&\rotatebox{90}{\tiny
QGRU}&\rotatebox{90}{\tiny
LSTM}&\rotatebox{90}{\tiny
QLSTM\phantom{.}}&\rotatebox{90}{\tiny
CNN}&\rotatebox{90}{\tiny
GRU}&\rotatebox{90}{\tiny
QGRU}&\rotatebox{90}{\tiny
LSTM}&\rotatebox{90}{\tiny
QLSTM}&\rotatebox{90}{\tiny CNN}&\rotatebox{90}{\tiny
GRU}&\rotatebox{90}{\tiny
QGRU}&\rotatebox{90}{\tiny
LSTM}&\rotatebox{90}{\tiny
QLSTM}&\rotatebox{90}{\tiny CNN}
&\rotatebox{90}{\tiny
GRU}&\rotatebox{90}{\tiny
QGRU}&\rotatebox{90}{\tiny
LSTM}&\rotatebox{90}{\tiny
QLSTM}
&\rotatebox{90}{\tiny
GRU}&\rotatebox{90}{\tiny
QGRU}&\rotatebox{90}{\tiny
LSTM}&\rotatebox{90}{\tiny
QLSTM}
&\rotatebox{90}{\tiny
GRU}&\rotatebox{90}{\tiny
QGRU}&\rotatebox{90}{\tiny
LSTM}&\rotatebox{90}{\tiny
QLSTM}
\\ \hline
$\mathrm{grad}_{1 s}^\mathrm{L2}$&
.61&.68&.67&.70&.68&
.45&.47&.25&.33&.79&
.26&.31&.07&.18&.74&
.48&.23&.63&.19&
.52&.27&.73&.22&
.09&.11&.19&.19 \\
$\mathrm{grad}_{1 p}^\mathrm{L2}$&
.57&.67&.67&.70&.74&
.40&.43&.26&.34&.70&
.18&.35&.07&.13&.66&
.48&.22&.63&.18&
.53&.26&.73&.21&
.09&.09&.18&.11 \\
$\mathrm{grad}_{\int s}^\mathrm{L2}$&
.71&.66&.69&.71&.70&
.58&.32&.26&.21&.82&
.23&.15&.11&.08&.76&
.69&.67&.68&.51&
.73&.70&.75&.55&
.19&.22&.20&.20 \\
$\mathrm{grad}_{\int p}^\mathrm{L2}$&
.71&.70&.72&.71&.77&
.56&.34&.30&.23&.81&
.13&.08&.14&.01&\gd{.78}&
.68&.77&.50&.70&
.74&.82&.54&.78&
.19&.21&.19&.30 \\
\rowcolor{gray!25}$\mathrm{grad}_{1 s}^\mathrm{dot}$&
.88&.85&.81&.77&.86&
.79&.76&.59&.72&\gd{.89}&
\gd{.80}&.70&.14&.47&\gd{.79}&
.81&.62&.73&.56&
.85&.66&.81&.59&
.42&.34&.46&.36 \\
\rowcolor{gray!25}$\mathrm{grad}_{1 p}^\mathrm{dot}$&
\underline{.92}&.88&.84&.79&\underline{.95}&
.78&.72&.59&.72&.81&
.71&.59&.20&.44&.69&
.79&.58&.74&.54&
.83&.61&.81&.56&
.41&.33&.46&.35 \\
\rowcolor{gray!25}$\mathrm{grad}_{\int s}^\mathrm{dot}$&
.84&\gd{.90}&.85&.87&.87&
\gd{.81}&.68&.60&.68&\gd{.89}&
\gd{.82}&.64&.21&.26&\gd{.80}&
\gd{.90}&\gd{.87}&.78&.84&
\gd{.94}&\gd{.92}&.83&.89&
\gd{.54}&.51&.46&.52 \\
\rowcolor{gray!25}$\mathrm{grad}_{\int p}^\mathrm{dot}$&
.86&\gd{.89}&.84&\gd{.89}&\bs{.96}&
\gd{.80}&.69&.62&.73&\gd{.89}&
\gd{.80}&.53&.40&.54&\gd{.78}&
\gd{.87}&\gd{.85}&.68&.84&
\gd{.93}&\gd{.92}&.74&\gd{.93}&
.53&.48&.42&.51 \\
$\mathrm{omit}_1$&
.79&.82&.85&.87&.61&
.78&.75&.54&.76&.82&
\gd{.80}&.48&.33&.48&.65&
.81&.81&.79&.80&
.86&.87&.86&.84&
.43&.45&.44&.45 \\
$\mathrm{omit}_3$&
\gd{.89}&.80&\gd{.89}&\gd{.88}&.59&
.79&.71&.72&\gd{.81}&.76&
.77&.37&.36&.49&.61&
.74&.77&.73&.73&
.82&.84&.82&.79&
.41&.45&.42&.46 \\
$\mathrm{omit}_7$&
\gd{.92}&.88&\gd{.91}&\gd{.91}&.70&
.79&.77&.77&\gd{.84}&.84&
.77&.49&.44&.55&.65&
.76&.80&.66&.74&
.85&.88&.78&.80&
.40&.48&.43&.47 \\
$\mathrm{occ}_1$&
.80&.71&.74&.84&.61&
.78&.73&.60&.77&.82&
.77&.49&.19&.10&.65&
\bs{.91}&\gd{.85}&\bs{.86}&\gd{.86}&
\gd{.94}&.88&\gd{.89}&.88&
.50&.44&.46&.47 \\
$\mathrm{occ}_3$&
\gd{.92}&.61&\gd{.93}&.85&.59&
.78&.63&.74&.74&.76&
.74&.37&.32&.35&.61&
.74&.73&.71&.72&
.78&.76&.76&.76&
.43&.37&.41&.43 \\
$\mathrm{occ}_7$&
\gd{.92}&.77&\gd{.93}&\gd{.90}&.70&
.78&.62&.74&.77&.84&
.74&.35&.43&.39&.65&
.64&.65&.63&.65&
.73&.73&.72&.73&
.36&.35&.39&.43 \\
\rowcolor{gray!25}$\mathrm{decomp}$&
.79&.88&\gd{.92}&\gd{.88}&-&
.75&.79&.77&.80&-&
.54&.36&\gd{.72}&.51&-&
.84&\gd{.87}&\bs{.86}&\gd{.90}&
\gd{.90}&\gd{.93}&\bs{.92}&\bs{.96}&
.52&\gd{.58}&\bs{.57}&\bs{.63} \\
$\mathrm{lrp}$&
\gd{.92}&.87&\gd{.91}&.84&.86&
\gd{.82}&\gd{.83}&\gd{.79}&\gd{.85}&\gd{.89}&
\bs{.85}&.72&\gd{.74}& \gd{.81}&\gd{.79}&
\gd{.90}&\bs{.90}&\bs{.86}&\bs{.91}&
\bs{.95}&\bs{.95}&\gd{.91}&\gd{.95}&
\gd{.58}&\bs{.60}&\gd{.52}&\bs{.63} \\
$\mathrm{deeplift}$&
\gd{.91}&\gd{.89}&\bs{.94}&.85&.87&
\gd{.82}&\gd{.83}&\gd{.78}&\gd{.84}&\gd{.89}&
\gd{.84}&.72&.70&\gd{.81}&\gd{.80}&
\bs{.91}&\bs{.90}&\gd{.85}&\bs{.91}&
\bs{.95}&\bs{.95}&\gd{.90}&\gd{.95}&
\bs{.59}&\gd{.59}&\gd{.52}&\bs{.63} \\
\rowcolor{gray!25}$\mathrm{limsse}^{\mathrm{bb}}$&
.81&.82&.83&.84&.78&
.78&.81&\gd{.78}&.80&.84&
.52&.53&.53&.54&.57&
.43&.41&.44&.42&
.54&.51&.56&.52&
.39&.43&.42&.41 \\
\rowcolor{gray!25}$\mathrm{limsse}^{\mathrm{ms}}_s$&
\bs{.94}&\bs{.94}&\gd{.93}&\bs{.93}&\gd{.91}&
\bs{.85}&\bs{.87}&\bs{.83}&\bs{.86}&\gd{.89}&
\bs{.85}&\bs{.84}&\bs{.76}&\bs{.84}&\bs{.82}&
.62&.62&.67&.63&
.75&.74&.82&.75&
.52&.53&\gd{.55}&.53 \\
\rowcolor{gray!25}$\mathrm{limsse}^{\mathrm{ms}}_p$&
.87&.88&.85&.86&\gd{.94}&
\bs{.85}&\gd{.86}&\bs{.83}&\bs{.86}&\bs{.90}&
\gd{.81}&\gd{.80}&\gd{.74}&.76&.76&
.62&.62&.67&.63&
.75&.74&.82&.75&
.51&.53&\gd{.55}&.53 \\ \hline
$\mathrm{random}$&
.69&.67&.70&.69&.66&
.20&.19&.22&.22&.21&
.09&.09&.06&.06&.08&
.27&.27&.27&.27&
.33&.33&.33&.33&
.12&.13&.12&.12 \\
$\mathrm{last}$&
-&-&-&-&-&-&-&-&-&-&-&-&-&-&-&
.66&.67&.66&.67&
.76&.77&.76&.77&
.21&.27&.25&.26 \\ \hline
$N$&
\multicolumn{5}{c|}{$7551\leq N \leq 7554$}&
\multicolumn{5}{c?}{$3022 \leq N \leq 3230$}&
\multicolumn{5}{c?}{$137 \leq N \leq 150$}&
\multicolumn{8}{c|}{$N \approx 1400000$}&
\multicolumn{4}{c}{$N \approx 20000$} \\
\end{tabular}
\caption{Pointing game accuracies in hybrid
	document experiment (left), on manually annotated benchmark (middle) and in
morphosyntactic agreement experiment (right). $\mathrm{hit}_\mathrm{target}$
(resp.\ $\mathrm{hit}_\mathrm{feat}$): maximal relevance on
subject (resp.\ on noun with the predicted number feature). Bold: top explanation method. Underlined: within 5 points of top explanation method.}
\label{tab:experiments-results}
}
\end{table*}

\section{Experiments}
\label{sec:experiments}

%% file: experiments-hybrid.tex
\subsection{Hybrid document experiment}
\label{sec:experiments-hybrid}
For the hybrid document experiment, we use the 20 newsgroups
corpus (topic classification) \cite{lang1995newsweeder} and reviews from the 10th yelp
dataset challenge (binary sentiment analysis)\footnote{\url{www.yelp.com/dataset\_challenge}}.
We train five DNNs per corpus: a bidirectional GRU \cite{cho2014properties}, a bidirectional LSTM \cite{hochreiter1997long}, a 1D CNN with global max pooling \cite{collobert2011natural}, a bidirectional Quasi-GRU (QGRU), and a bidirectional Quasi-LSTM (QLSTM).
The Quasi-RNNs are 1D CNNs with a feature-wise gated recursive pooling layer \cite{bradbury2016quasi}.
Word embeddings are $\mathbb{R}^{300}$ and initialized with pre-trained GloVe embeddings \cite{pennington2014glove}\footnote{\url{http://nlp.stanford.edu/data/glove.840B.300d.zip}}.
The main layer has a hidden size of 150 (bidirectional architectures: 75 dimensions per direction).
For the QRNNs and CNN, we use a kernel width of 5.
In all five architectures, the resulting document representation is projected to 20 (resp. two) dimensions using a fully connected layer, followed by a softmax.
See supplementary material for details on training and regularization.

After training, we sentence-tokenize the test sets, shuffle the sentences, concatenate ten sentences at a time and classify the resulting hybrid documents.
Documents that are assigned a class that is not the gold label of at least one constituent word are discarded (yelp: $<$ 0.1\%; 20 newsgroups: 14\% - 20\%).
On the remaining documents, we use the explanation methods from \secref{explanation-methods} to find the maximally relevant word for each prediction.
The random baseline samples the maximally relevant word from a uniform distribution.

For reference, we also evaluate on a
\textbf{human judgment} benchmark (\citet{mohseni2018human}, \tabref{experiments-results}, C11-C15).
It contains 188 documents from the 20 newsgroups test set (classes sci.med and sci.electronics), with one manually created list of relevant words per document.
We discard documents that are incorrectly classified (20\% - 27\%) and define: $\mathrm{hit}(\phi, \mathbf{X}) = \mathbb{I}[\mathrm{rmax}(\mathbf{X}, \phi) \in \mathrm{gt}(\mathbf{X})]$, 
where $\mathrm{gt}(\mathbf{X})$ is the manual ground truth.

%% file: experiments-morphosyntactic.tex
\subsection{Morphosyntactic agreement experiment}
\label{sec:experiments-morphosyntactic}
For the morphosyntactic agreement experiment, we use automatically annotated English Wikipedia sentences by \citet{linzen2016assessing}\footnote{\url{www.tallinzen.net/media/rnn\_agreement/agr\_50\_mostcommon\_10K.tsv.gz}}.
For our purpose, a sample consists of:
all words preceding the verb: $\mathbf{X} = [x_1 \cdots x_{T}]$;
part-of-speech (POS) tags: $\mathrm{pos}(\mathbf{X}, t)$ $\in$ $\{$VBZ, VBP, NN, NNS, $\ldots\}$; and
the position of the subject: $\mathrm{target}(\mathbf{X}) \in [1,T]$.
The number feature is derived from the POS:
\begin{equation}
\mathrm{feat}(\mathbf{X}, t) = 
\begin{cases}
\text{Sg} & \text{ if } \mathrm{pos}(\mathbf{X},t) \in \{\text{VBZ, NN}\} \\
\text{Pl} & \text{ if } \mathrm{pos}(\mathbf{X},t) \in \{\text{VBP, NNS}\} \\
\text{n/a} & \text { otherwise} 
\end{cases}\nonumber
\end{equation}
The gold label of a sentence is the number of its verb, i.e., $y(\mathbf{X}) = \mathrm{feat}(\mathbf{X}, T+1)$.

As task methods, we replicate
\citet{linzen2016assessing}'s unidirectional LSTM ($\mathbb{R}^{50}$ randomly initialized word embeddings, hidden size 50).
We also train unidirectional GRU, QGRU and QLSTM architectures with the same dimensionality.
We use the explanation methods from \secref{explanation-methods} to find the most relevant word for predictions on the test set.
As described in \secref{evaluation-paradigms-morphosyntactic}, explanation methods are awarded a $\mathrm{hit}_\mathrm{target}$ (resp. $\mathrm{hit}_\mathrm{feat}$) point if this word is the subject (resp. a noun with the predicted number feature).
For reference, we use a random baseline as well as a baseline that assumes that the most relevant word directly precedes the verb.

%% file: example-hybrid-limsse.tex
\begin{figure*}
\centering
\scriptsize
\begin{tabular}{|m{0.08\textwidth}m{0.88\textwidth}|}
\hline
${\mathrm{decomp}}$&\input{examples/LSTM_gamma_2167} [is ...]\\
${\mathrm{deeplift}}$&\input{examples/LSTM_deeplift_2167} [is ...]\\
${\mathrm{limsse}^\mathrm{ms}_p}$&\input{examples/LSTM_lime_prob_2167} [is ...]\\
\hline \hline 
${{\mathrm{lrp}}}$& \input{examples/QLSTM_yelp_lrp_2089.tex} \\
${{\mathrm{limsse}^\mathrm{ms}_p}}$& \input{examples/QLSTM_yelp_lime_prob_2089.tex} \\ \hline
\end{tabular}
\caption{\textbf{Top:} verb context classified {singular}. Task method: LSTM.
\textbf{Bottom:} hybrid yelp review, classified {positive}. Task method:
  QLSTM.}
\label{fig:example-hybrid-limsse}
\label{fig:example-morphosyntactic-limsse}
\end{figure*}

%% file: discussion.tex
\section{Discussion}
\label{sec:discussion}
\subsection{Explanation methods}
Our experiments suggest that explanation methods for neural NLP differ in quality.

As in previous work (see \secref{related-work}),
\textbf{gradient L2 norm} ($\mathrm{grad}^\mathrm{L2}$)
performs poorly, especially on RNNs.
We assume that this is due to its inability to distinguish relevances for and against $k$.

\textbf{Gradient embedding dot product}
($\mathrm{grad}^\mathrm{dot}$) is competitive on
CNN (\tabref{experiments-results},
$\mathrm{grad}^\mathrm{dot}_{1 p}$ C05,
$\mathrm{grad}^\mathrm{dot}_{1 s}$ C10, C15), presumably
because $\mathrm{relu}$ is
linear on positive inputs, so gradients are exact instead of
approximate.  $\mathrm{grad}^\mathrm{dot}$
also has decent performance for GRU ($\mathrm{grad}^\mathrm{dot}_{1 p}$ C01, $\mathrm{grad}^\mathrm{dot}_{\int s}$ C\{06, 11, 16, 20,
24\}), perhaps because GRU hidden
activations are always in [-1,1], where $\tanh$ and $\sigma$ are approximately linear.

\textbf{Integrated gradient} ($\mathrm{grad}_{\int}$) mostly
outperforms simple gradient ($\mathrm{grad}_1$), though not
consistently (C01, C07).  Contrary to expectation, integration did not
help much with the failure of the gradient method on LSTM on 
20 newsgroups ($\mathrm{grad}^\mathrm{dot}_{1}$ vs.
$\mathrm{grad}^\mathrm{dot}_{\int}$ in C08, C13),
which we had assumed to be due to saturation of $\tanh$ on
large absolute activations in $\vec{c}$.
Smaller intervals may be needed to approximate the integration,
however, this means additional computational cost.

The gradient of $s(k, \mathbf{X})$ performs
better or similar to the gradient of
$p(k|\mathbf{X})$. The main exception is yelp
($\mathrm{grad}^\mathrm{dot}_{1 s}$ vs.\
$\mathrm{grad}^\mathrm{dot}_{1 p}$, C01-C05).
This is probably due to conflation by $p(k|\mathbf{X})$ of evidence
for $k$ (numerator in \eqref{softmaxp}) and against
competitor classes (denominator). In a
two-class scenario, there is little incentive to keep
classes separate, leading to information flow through
the denominator.  In future work, we will replace the two-way
softmax with a one-way sigmoid such that $\phi(t,0,\mathbf{X}) := -\phi(t,1,\mathbf{X})$.

\textbf{LRP} and \textbf{DeepLIFT} are the most consistent
explanation methods across evaluation paradigms and task
methods. (The comparatively low pointing game accuracies on the
yelp QRNNs and CNN (C02, C04, C05) are probably due to
the fact that they explain $s(k, .)$
in a two-way softmax, see above.) On CNN (C05, C10,
C15), LRP and $\mathrm{grad}^\mathrm{dot}_{1 s}$ perform
almost identically, suggesting that they are indeed
quasi-equivalent on this architecture (see
\secref{explanation-methods-lrp}).  On (Q)RNNs,
modified LRP and DeepLIFT appear to be superior to the
gradient method ($\mathrm{lrp}$ vs.\
$\mathrm{grad}^\mathrm{dot}_{1 s}$,
$\mathrm{deeplift}$ vs.\
$\mathrm{grad}^\mathrm{dot}_{\int s}$, C01-C04,
C06-C09, C11-C14, C16-C27).

\textbf{Decomposition} performs well on LSTM, 
especially in the morphosyntactic agreement experiment,
but it is inconsistent on other architectures.
Gated RNNs have a long-term additive and a
multiplicative pathway, and the decomposition method
only detects information traveling via the additive one.
\citet{miao2016simplifying} show qualitatively that GRUs
often reorganize long-term memory abruptly, which might
explain the difference between LSTM and GRU. QRNNs only have
additive recurrent connections; however, given that
$\vec{c}_t$ (resp. $\vec{h}_t$) is calculated by convolution
over several time steps, decomposition relevance can be
incorrectly attributed inside that window.  
This likely is the reason for the stark difference
between the performance of decomposition on QRNNs in the
hybrid document experiment and on the manually labeled data
(C07, C09 vs.\ C12, C14).
Overall, we do not recommend the decomposition method, because
it fails to take into account all routes by
which information can be propagated.

\textbf{Omission and occlusion} produce inconsistent results in the hybrid document experiment.
\citet{shrikumar2017learning} show that perturbation methods
can lack sensitivity when there are more relevant
inputs than the ``perturbation window'' covers.
In the morphosyntactic agreement experiment, omission is not competitive; we assume that this is because it interferes 
too much with syntactic structure.
$\mathrm{occ}_1$ does better (esp. C16-C19), possibly because an all-zero ``placeholder'' is less disruptive than word removal.
But despite some high scores, it is less consistent than other explanation methods.

Magnitude-sensitive \textbf{LIMSSE}
(${\mathrm{limsse}^\mathrm{ms}}$) consistently outperforms
black-box LIMSSE (${\mathrm{limsse}^\mathrm{bb}}$), which
suggests that numerical outputs should be used for
approximation where possible.  In the hybrid document experiment,
magnitude-sensitive LIMSSE outperforms the other explanation
methods (exceptions: C03, C05).  However, it fails in the morphosyntactic agreement experiment (C16-C27). 
In fact, we expect LIMSSE to be unsuited for
\textit{large context} problems, as it cannot discover dependencies
whose range is bigger than a given text sample.
In \figref{example-morphosyntactic-limsse} (top), $\mathrm{limsse}^\mathrm{ms}_p$ highlights \textit{any} singular
noun without taking into account how that noun fits into the overall syntactic structure.

\subsection{Evaluation paradigms}
The assumptions made by our automatic evaluation paradigms
have exceptions: (i) the correlation between fragment 
of origin and relevance does not always hold (e.g.,
a positive review may contain negative fragments,
and will almost certainly contain neutral fragments); 
(ii) in morphological prediction, we cannot always expect the subject to be the only
predictor for number.  In
\figref{example-morphosyntactic-snippet} (bottom) for example,
``few'' is a reasonable clue for plural despite not being a noun.
This imperfect ground truth means that absolute pointing
game accuracies should be taken with a grain of salt; but we argue that
this does not invalidate them for comparisons.

We also point out that there are characteristics of
explanations that may be desirable but are not reflected by the
pointing game.  Consider \figref{example-hybrid-limsse} (bottom).  Both explanations
get hit points, but the $\mathrm{lrp}$
explanation appears ``cleaner'' than
$\mathrm{limsse}^\mathrm{ms}_p$, with relevance
concentrated on fewer tokens.

%% file: related-work.tex
\section{Related work}
\label{sec:related-work}
\subsection{Explanation methods}
Explanation methods can be divided into local and global methods \cite{doshi2017towards}.
Global methods infer general statements about what a
DNN has learned, e.g., by 
clustering documents \cite{aubakirova2016interpreting} or n-grams \cite{kadar2017representation} according to the neurons that they activate.
\citet{li2015visualizing} compare embeddings of specific words with reference points to measure how drastically they were changed during training.
In computer vision, \citet{simonyan2013deep} optimize the input space to maximize the activation of a specific neuron.
Global explanation methods are of limited value for
explaining a specific prediction as they represent average behavior. 
Therefore, we focus on local methods.

Local explanation methods explain a decision taken for one
specific input at a time.
We have attempted to include all important
local methods for NLP in our experiments
(see \secref{explanation-methods}).
We do not address self-explanatory models (e.g., attention \cite{bahdanau2014neural} or rationale models \cite{lei2016rationalizing}),
as these are very specific architectures that may not be not applicable to all tasks.

\subsection{Explanation evaluation}
According to \citet{doshi2017towards}'s taxonomy of explanation evaluation paradigms,
\emph{application-grounded} paradigms test how well an explanation method helps real users solve real tasks (e.g., doctors judge automatic diagnoses);
\emph{human-grounded} paradigms rely on proxy tasks (e.g., humans rank task methods based on explanations);
\emph{functionally-grounded} paradigms work without human input, like our approach.

\citet{arras2016explaining} (cf. \citet{samek2016evaluating}) propose a functionally-grounded explanation evaluation paradigm for NLP where words in a correctly (resp. incorrectly) classified document are deleted in descending (resp. ascending) order of relevance.
They assume that the fewer words must be deleted to reduce (resp. increase) accuracy, the better the explanations.
According to this metric, LRP (\secref{explanation-methods-lrp}) outperforms $\mathrm{grad}^\mathrm{L2}$ on CNNs \cite{arras2016explaining} and LSTMs \cite{arras2017explaining} on 20 newsgroups.
\citet{ancona2017unified} perform the same experiment with a binary sentiment analysis LSTM.
Their graph shows $\mathrm{occ}_1$, $\mathrm{grad}^{\mathrm{dot}}_1$ and $\mathrm{grad}^{\mathrm{dot}}_{\int}$ tied in first place, while LRP, DeepLIFT and the gradient L1 norm lag behind.
Note that their treatment of LSTM gates in LRP / DeepLIFT differs from our implementation.

An issue with the word deletion paradigm is that it uses syntactically broken inputs, which may introduce artefacts \cite{sundararajan2017axiomatic}.
In our hybrid document paradigm, inputs are syntactically intact (though semantically incoherent at the document level); the morphosyntactic agreement paradigm uses unmodified inputs.

Another class of functionally-grounded evaluation paradigms interprets the performance of a secondary task method, on inputs that are derived from (or altered by) an explanation method, as a proxy for the quality of that explanation method.
\citet{murdoch2017automatic} build a rule-based classifier from the most relevant phrases in a corpus (task method: LSTM).
The classifier based on $\mathrm{decomp}$ (\secref{explanation-methods-gamma}) outperforms the gradient-based classifier, which is in line with our results.
\citet{arras2016relevant} build document representations by summing over word embeddings weighted by relevance scores (task method: CNN).
They show that K-nearest neighbor performs better on document representations derived with LRP than on those derived with $\mathrm{grad}^\mathrm{L2}$, which also matches our results.
\citet{denil2014extraction} condense documents by extracting top-K relevant sentences, and let the original task method (CNN) classify them.
The accuracy loss, relative to uncondensed documents, is smaller for $\mathrm{grad}^\mathrm{dot}$ than for heuristic baselines.

In the domain of human-based evaluation paradigms, \citet{ribeiro2016should} compare different variants of LIME (\secref{explanation-methods-lime}) by how well they help non-experts clean a corpus from words that lead to overfitting.
\citet{selvaraju2017grad} assess how well explanation methods help non-experts identify the more accurate out of two object recognition CNNs.
These experiments come closer to real use cases than functionally-grounded paradigms; however, they are less scalable.

%% file: summary.tex
\section{Summary}
\label{sec:summary}
We conducted the first comprehensive evaluation of
explanation methods for NLP, an important undertaking
because there is a need for understanding the behavior
of DNNs.

To conduct this study, we introduced evaluation paradigms for explanation
methods for two classes of NLP tasks, small context tasks
(e.g., topic classification) and large context tasks
(e.g., morphological prediction).
Neither paradigm requires manual
annotations.
We also introduced LIMSSE, a substring-based explanation
method inspired by LIME and designed for NLP.

Based on our experimental results, we recommend LRP,
DeepLIFT and LIMSSE for small context tasks and LRP and
DeepLIFT for large context tasks, on all five DNN architectures that we tested.
On CNNs and possibly
GRUs, the (integrated) gradient embedding dot product is a
good alternative to DeepLIFT and LRP.

%% file: explanation-methods-implementation.tex
\section{Code}
Our implementation of LIMSSE, the gradient, perturbation and decomposition methods can be found in our branch of the \texttt{keras} package: \url{www.github.com/NPoe/keras}.
To re-run our experiments, see scripts in \url{www.github.com/NPoe/neural-nlp-explanation-experiment}.
Our LRP implementation (same repository) is adapted from \citet{arras2017explaining}\footnote{\url{https://github.com/ArrasL/LRP_for_LSTM}}.

%% file: appendix.tex
\small
\section{Corpora and data preprocessing}
The 20 newsgroups corpus \cite{lang1995newsweeder} was downloaded using the Python \texttt{sklearn}
package \cite{pedregosa2011scikit}, removing all headers, footers and quotes.
The corpus contains 18,846 posts and comes with a training and test set.
We randomly split the latter into a heldout and a test set. 

For sentiment analysis we use the Pennsylvania subset of the 10th yelp dataset challenge\footnote{\url{www.yelp.com/dataset\_challenge}}.
It contains 206,338 reviews with 1 to 5 star ratings. 
1 or 2 stars are mapped to ``negative'', 4 or 5 stars to ``positive'', 3 star
reviews are discarded.  
We randomly split the data into training, heldout and test sets (90\%/5\%/5\%). 
On both corpora, we use NLTK \cite{bird2009natural} for word and sentence
tokenization. 
Words with a frequency rank above 50000 are mapped to \textit{oov}.
To create hybrid documents, we sentence-tokenize the test sets, shuffle, and then concatenate ten sentences at a time.

The manually annotated 20 newsgroups documents were obtained from \citet{mohseni2018human}\footnote{\url{http://github.com/SinaMohseni/ML-Interpretability-Evaluation-Benchmark}}.
The relevance ground truth consists of one list of lowercased word types per document.
There are a number of mismatches between the ground truth and the documents (e.g., one list contains \textit{rays} but its document only contains \textit{x-rays}).
This made some reverse engineering necessary:
Given $\mathbf{X}$ and its list, we add $t$ to $\mathrm{gt}(\mathbf{X})$ if lower-cased $x_t$ is a prefix or suffix of at least one word type in the list.

For the morphosyntactic agreement experiment, we use
\citet{linzen2016assessing}'s
corpus of 1,577,211 English Wikipedia sentences
with automatic morphosyntactic annotation\footnote{\url{www.tallinzen.net/media/rnn\_agreement/agr\_50\_mostcommon\_10K.tsv.gz}}.  
We replicate the
original dataset sizes (9\% train, 1\% heldout, 90\% test).
Like in the original corpus, words with a frequency rank above 10,000 are replaced by their part-of-speech tag.

\section{Neural networks}
Every neural network used in our paper is made up of a word embedding matrix, followed by a core layer, followed by a fully-connected layer with softmax activation.

In the hybrid document experiment, the $|V|\times 300$ embedding matrix is initialized with GloVe embeddings \cite{pennington2014glove}\footnote{\url{http://nlp.stanford.edu/data/glove.840B.300d.zip}}, which are fine-tuned during training.
The core layer is a bidirectional Gated Recurrent Unit (GRU, \citet{cho2014properties}), bidirectional Long-Short Term Memory Network (LSTM, \citet{hochreiter1997long}), bidirectional Quasi-GRU or Quasi-LSTM \cite{bradbury2016quasi}, or a 1D Convolutional Neural Network (CNN) with global max pooling \cite{collobert2011natural}.
In all cases, the core layer has a hidden size of 150 (bidirectional architectures: 75 per direction), for QRNNs and CNN, we use a kernel width of 5.
For regularization, we use 50\% dropout between layers and on hidden-to-hidden connections (GRU/LSTM only).

We minimize categorical crossentropy using Adam \cite{kingma2014adam}, with learning rate 0.001, $\beta_1=0.9$, $\beta_2=0.999$ and batch size 8.
Heldout accuracy is monitored; after two stagnant epochs, the learning rate is halved, and after 5 (yelp), resp. 25 (20 newsgroups), stagnant epochs, training is stopped and the model from the best epoch is stored.
Final test set accuracies are .964/.954/.965/.959/.957 on yelp and .727/.716/.730/.735/.705 on 20 newsgroups (GRU/QGRU/LSTM/QLSTM/CNN).

In the morphosyntactic agreement experiment, the $|V|\times 50$ embedding matrix is randomly initialized.
All (Q)RNNs are unidirectional and have a hidden size of 50.
QRNN kernel width is 5.
The core layer is followed by a fully connected $50 \times 2$ layer with softmax activation.
We minimize categorical crossentropy using Adam (see above), with early stopping after 20 epochs based on heldout accuracy, and a batch size of 16.
Final test set accuracies are .991/.985/.990/.986 (GRU/QGRU/LSTM/QLSTM).
Contrary to \citet{linzen2016assessing}, we do not train an ensemble.

\subsection{GRU}
\begin{align}
\displaystyle
		\nonumber
        \vec{h}_0 & = 0 \\ \nonumber
        \vec{z}_t & = \sigma(\mathbf{V_z}  \vec{e}_t + \mathbf{U_z}  \vec{h}_{t-1} + \vec{b}_z) \\ \nonumber
        \vec{r}_t & = \sigma(\mathbf{V_r}  \vec{e}_t + \mathbf{U_r}  \vec{h}_{t-1} + \vec{b}_r) \\ \nonumber
        \vec{g'}_t & = \mathbf{V}  \vec{e}_t + \mathbf{U}  (\vec{r}_t \odot \vec{h}_{t-1}) + \vec{b} \\ \nonumber
        \vec{g}_t &= \mathrm{tanh}(\vec{g'}_t) \\ \nonumber
        \vec{h}_t & = \vec{z}_t \odot \vec{h}_{t-1} + (\vec{1}-\vec{z}_t) \odot \vec{g}_t \\ \nonumber
\end{align}
\subsection{QGRU}
\begin{align}
\displaystyle
\nonumber
\mathbf{Z} & = \sigma(\mathbf{V_z} \star [0 \ldots \vec{e}_1 \ldots \vec{e}_T] + \vec{b}_z) \\ \nonumber
\mathbf{G'} & = \mathbf{V} \star [0 \ldots \vec{e}_1 \ldots \vec{e}_T] + \vec{b} \\ \nonumber
\mathbf{G} &= \mathrm{tanh}(\mathbf{G'}) \\ \nonumber
\vec{h}_0 &= 0 \\ \nonumber
\vec{h}_t &= \vec{z}_t \odot \vec{h}_{t-1} + (1-\vec{z}_t) \odot \vec{g}_t \\ \nonumber
\end{align}
\subsection{LSTM}
\begin{align}
\displaystyle
		\nonumber
        \vec{c}_0 & = \vec{h}_0 = 0 \\ \nonumber
        \vec{i}_t & = \sigma(\mathbf{V_i}  \vec{e}_t + \mathbf{U_i}  \vec{h}_{t-1} + \vec{b}_i) \\ \nonumber
        \vec{f}_t & = \sigma(\mathbf{V_f}  \vec{e}_t + \mathbf{U_f}  \vec{h}_{t-1} + \vec{b}_f) \\ \nonumber
        \vec{o}_t & = \sigma(\mathbf{V_o}  \vec{e}_t + \mathbf{U_o}  \vec{h}_{t-1} + \vec{b}_o) \\ \nonumber
        \vec{g'}_t & = \mathbf{V}  \vec{e}_t + \mathbf{U}  \vec{h}_{t-1} + \vec{b} \\ \nonumber
        \vec{g}_t &= \mathrm{tanh}(\vec{g'}_t) \\ \nonumber
        \vec{c}_t & = \vec{f}_t \odot \vec{c}_{t-1} + \vec{i}_t \odot \vec{g}_t \\ \nonumber
        \vec{h}_t & = \vec{o}_t \odot \mathrm{tanh}(\vec{c}_t) \\ \nonumber
\end{align}
\subsection{QLSTM}
\begin{align}
\displaystyle
\nonumber
\mathbf{I} & = \sigma(\mathbf{V_i} \star [0 \ldots \vec{e}_1 \ldots \vec{e}_T] + \vec{b}_i) \\ \nonumber
\mathbf{F} & = \sigma(\mathbf{V_f} \star [0 \ldots \vec{e}_1 \ldots \vec{e}_T] + \vec{b}_f) \\ \nonumber
\mathbf{O} & = \sigma(\mathbf{V_o} \star [0 \ldots \vec{e}_1 \ldots \vec{e}_T] + \vec{b}_o) \\ \nonumber
\mathbf{G'} & = \mathbf{V} \star [0 \ldots \vec{e}_1 \ldots \vec{e}_T] + \vec{b} \\ \nonumber
\mathbf{G} &= \mathrm{tanh}(\mathbf{G'}) \\ \nonumber
\vec{h}_0 &= \vec{c}_0 = 0 \\ \nonumber
\vec{c}_t &= \vec{f}_t \odot \vec{c}_{t-1} + \vec{i}_t \odot \vec{g}_t \\ \nonumber
\vec{h}_t &= \vec{o}_t \odot \mathrm{tanh}(\vec{c}_t) \\ \nonumber
\end{align}
\subsection{CNN}
\begin{align}
\displaystyle
\nonumber
\mathbf{G'} & = \mathbf{V} \star [0 \ldots \vec{e}_1 \ldots \vec{e}_T \ldots 0] + \vec{b} \\ \nonumber
\mathbf{G} &= \mathrm{relu}(\mathbf{G'}) \\ \nonumber
h_d &= \underset{t}{\mathrm{max}}(g_{t,d}) \\ \nonumber
\end{align}

\section{RGB coding in examples}
\begin{align}
\label{eq:visualize}
\nonumber
\phi'(t,k,\mathbf{X})& = \frac{\phi(t,k,\mathbf{X})}{\mathrm{max}_{t'}(1.1 |\phi(t',k,\mathbf{X})|)} \\ \nonumber
R(t,k,\mathbf{X}) & = \phi'(t,k,\mathbf{X}) \mathbb{I}[\phi(t,k,\mathbf{X}) < 0] \\ \nonumber
G(t,k,\mathbf{X}) & = \phi'(t,k,\mathbf{X}) \mathbb{I}[\phi(t,k,\mathbf{X}) > 0] \\ \nonumber
B(t,k,\mathbf{X}) & = 0\\ \nonumber
\end{align}

\section{Epsilon LRP and DeepLIFT}
In the following, we assume that the hidden layer relevance vector $R(\vec{h})$ (resp. $R(\vec{h}_T)$) has been backpropagated by the upstream fully connected layer using equations from Sections 3.2 and 3.3 (main paper).
DeepLIFT can be derived by replacing $h$, $g$, $g'$, $e$, $c$ with $h-\bar{h}$, $g-\bar{g}$, $g'-\bar{g}'$, $e-\bar{e}$, $c-\bar{c}$.
F is CNN / QRNN kernel width.
\subsection{GRU}
\begin{align}
\displaystyle
\nonumber
R(g_{t,d}) & = R(h_{t,d}) \frac{g_{t,d} \cdot (1-z_{t,d}) }{h_{t,d} + \mathrm{esign}(h_{t,d})} \\ \nonumber
R(e_{t,d}) & = \sum_{j=1}^{\mathrm{dim}(\vec{g}_t)} R(g_{t,j})\frac{ e_{t,d} \cdot v_{d,j}}{g'_{t,j} + \mathrm{esign}(g'_{t,j})} \\ \nonumber
R(h_{t-1,d}) & = R(h_{t,d}) \frac{h_{t-1, d} \cdot z_{t,d}}{h_{t,d} + \mathrm{esign}(h_{t,d})} \\ \nonumber
& + \sum_{j=1}^{\mathrm{dim}(\vec{g}_t)} R(g_{t,j})\frac{ h_{t-1,d} \cdot r_{t,d} \cdot u_{d,j}}{g'_{t,j} + \mathrm{esign}(g'_{t,j})} \\ \nonumber
\end{align}
\subsection{QGRU}
\begin{align}
\displaystyle
\nonumber
R(g_{t,d}) & = R(h_{t,d})  \frac{g_{t,d} \cdot (1-z_{t,d}) }{h_{t,d} + \mathrm{esign}(h_{t,d})} \\ \nonumber
R(h_{t-1,d}) & = R(h_{t,d})\frac{ h_{t-1,d} \cdot z_{t,d}}{h_{t,d} + \mathrm{esign}(h_{t,d})} \\ \nonumber
R(e_{t,d}) & = \sum_{j=1}^{\mathrm{dim}(\vec{g}_t)} \sum_{k=0}^{F-1} R(g_{t+k,j})\frac{ e_{t,d} \cdot v_{k,d,j}}{g'_{t+k,j} + \mathrm{esign}(g'_{t+k,j})}
\end{align}
\subsection{LSTM}
\begin{align}
\displaystyle
\nonumber
R(c_{T+1,d}) & = 0 \\ \nonumber
R(c_{t,d}) & =R(h_{t,d}) \frac{\mathrm{tanh}(c_{t,d}) \cdot o_{t,d} }{h_{t,d} + \mathrm{esign}(h_{t,d})} \\ \nonumber
& + R(c_{t+1,d}) \frac{c_{t,d} \cdot f_{t+1,d} }{c_{t+1,d} + \mathrm{esign}(c_{t+1,d})} \\ \nonumber
R(g_{t,d}) & = R(c_{t,d})  \frac{g_{t,d} \cdot i_{t,d} }{c_{t,d} + \mathrm{esign}(c_{t,d})} \\ \nonumber
R(e_{t,d}) & = \sum_{j=1}^{\mathrm{dim}(\vec{g}_t)} R(g_{t,j}) \frac{e_{t,d} \cdot v_{d,j}}{g'_{t,j} + \mathrm{esign}(g'_{t,j})} \\ \nonumber
R(h_{t-1,d}) & = \sum_{j=1}^{\mathrm{dim}(\vec{g}_t)} R(g_{t,j}) \frac{h_{t-1,d} \cdot u_{d,j}}{g'_{t,j} + \mathrm{esign}(g'_{t,j})} \\ \nonumber
\end{align}
\subsection{QLSTM}
\begin{align}
\displaystyle
\nonumber
R(c_{T+1,d}) &= 0 \\ \nonumber
R(c_{t,d}) & = R(h_{t,d}) \frac{\mathrm{tanh}(c_{t,d}) \cdot o_{t,d} }{h_{t,d} + \mathrm{esign}(h_{t,d})} \\ \nonumber
& +R(c_{t+1,d})  \frac{c_{t,d} \cdot f_{t+1,d} }{c_{t+1,d} + \mathrm{esign}(c_{t+1,d})} \\ \nonumber
R(g_{t,d}) & = R(c_{t,d})\frac{g_{t,d} \cdot i_{t,d} }{c_{t,d} + \mathrm{esign}(c_{t,d})} \\ \nonumber
R(e_{t,d}) & = \sum_{j=1}^{\mathrm{dim}(\vec{g}_t)}  \sum_{k=0}^{F-1} R(g_{t+k,j}) \frac{ e_{t,d} \cdot v_{k,d,j}}{g'_{t+k,j} + \mathrm{esign}(g'_{t+k,j})}
\end{align}
\subsection{CNN}
\begin{align}
\displaystyle
\nonumber
F' &= \frac{F-1}{2} \\ \nonumber
R(g_{t,d}) &= R(h_d) \cdot \mathbb{I}[\mathrm{argmax}_{t'}(g_{t',d}) = t] \\ \nonumber
R(e_{t,d}) & = \sum_{j=1}^{\mathrm{dim}(\vec{g})} \sum_{k=-F'}^{F'}  R(g_{t+k,j}) \frac{ e_{t,d} \cdot v_{k,d,j}}{g'_{t+k,j} + \mathrm{esign}(g'_{t+k,j})}
\end{align}

%% file: example-morphosyntactic1.tex
\begin{figure}[b!]
\centering
\tiny
\begin{tabular}{lll}
	${\mathrm{grad}^\mathrm{L2}_{1 s}}$&\input{examples/GRU_l2_raw_2109} [are ...]&\multirow{20}{*}{\rot{GRU}}\\
\gr${\mathrm{grad}^\mathrm{L2}_{1 p}}$&\input{examples/GRU_l2_prob_2109} [are ...]\\
${\mathrm{grad}^\mathrm{L2}_{\int s}}$&\input{examples/GRU_l2_raw_integrated_2109} [are ...]\\
\gr${\mathrm{grad}^\mathrm{L2}_{\int p}}$&\input{examples/GRU_l2_prob_integrated_2109} [are ...]\\
${\mathrm{grad}^\mathrm{dot}_{1 s}}$&\input{examples/GRU_salience_raw_2109} [are ...]\\
\gr${\mathrm{grad}^\mathrm{dot}_{1 p}}$&\input{examples/GRU_salience_prob_2109} [are ...] \\
${\mathrm{grad}^\mathrm{dot}_{\int s}}$&\input{examples/GRU_salience_raw_integrated_2109} [are ...]\\
\gr${\mathrm{grad}^\mathrm{dot}_{\int p}}$&\input{examples/GRU_salience_prob_integrated_2109} [are ...] \\
${\mathrm{omit}_1}$&\input{examples/GRU_omission_2109} [are ...]\\
\gr${\mathrm{omit}_3}$&\input{examples/GRU_omission-3_2109} [are ...]\\
${\mathrm{omit}_7}$&\input{examples/GRU_omission-7_2109} [are ...]\\
\gr${\mathrm{occ}_1}$&\input{examples/GRU_occlusion_2109} [are ...]\\
${\mathrm{occ}_3}$&\input{examples/GRU_occlusion-3_2109} [are ...]\\
\gr${\mathrm{occ}_7}$&\input{examples/GRU_occlusion-7_2109} [are ...]\\
${\mathrm{decomp}}$&\input{examples/GRU_gamma_2109} [are ...]\\
\gr${\mathrm{lrp}}$&\input{examples/GRU_lrp_2109} [are ...]\\
${\mathrm{deeplift}}$&\input{examples/GRU_deeplift_2109} [are ...]\\
\gr${\mathrm{limsse}^\mathrm{bb}}$&\input{examples/GRU_lime_class_2109} [are ...]\\
${\mathrm{limsse}^\mathrm{ms}_s}$&\input{examples/GRU_lime_raw_2109} [are ...]\\
\gr${\mathrm{limsse}^\mathrm{ms}_p}$&\input{examples/GRU_lime_prob_2109} [are ...]\\ \cline{1-2}
${\mathrm{grad}^\mathrm{L2}_{1 s}}$&\input{examples/QGRU_l2_raw_2109} [are ...]&\multirow{20}{*}{\rot{QGRU}}\\
\gr${\mathrm{grad}^\mathrm{L2}_{1 p}}$&\input{examples/QGRU_l2_prob_2109} [are ...]\\
${\mathrm{grad}^\mathrm{L2}_{\int s}}$&\input{examples/QGRU_l2_raw_integrated_2109} [are ...]\\
\gr${\mathrm{grad}^\mathrm{L2}_{\int p}}$&\input{examples/QGRU_l2_prob_integrated_2109} [are ...]\\
${\mathrm{grad}^\mathrm{dot}_{1 s}}$&\input{examples/QGRU_salience_raw_2109} [are ...]\\
\gr${\mathrm{grad}^\mathrm{dot}_{1 p}}$&\input{examples/QGRU_salience_prob_2109} [are ...] \\
${\mathrm{grad}^\mathrm{dot}_{\int s}}$&\input{examples/QGRU_salience_raw_integrated_2109} [are ...]\\
\gr${\mathrm{grad}^\mathrm{dot}_{\int p}}$&\input{examples/QGRU_salience_prob_integrated_2109} [are ...] \\
${\mathrm{omit}_1}$&\input{examples/QGRU_omission_2109} [are ...]\\
\gr${\mathrm{omit}_3}$&\input{examples/QGRU_omission-3_2109} [are ...]\\
${\mathrm{omit}_7}$&\input{examples/QGRU_omission-7_2109} [are ...]\\
\gr${\mathrm{occ}_1}$&\input{examples/QGRU_occlusion_2109} [are ...]\\
${\mathrm{occ}_3}$&\input{examples/QGRU_occlusion-3_2109} [are ...]\\
\gr${\mathrm{occ}_7}$&\input{examples/QGRU_occlusion-7_2109} [are ...]\\
${\mathrm{decomp}}$&\input{examples/QGRU_gamma_2109} [are ...]\\
\gr${\mathrm{lrp}}$&\input{examples/QGRU_lrp_2109} [are ...]\\
${\mathrm{deeplift}}$&\input{examples/QGRU_deeplift_2109} [are ...]\\
\gr${\mathrm{limsse}^\mathrm{bb}}$&\input{examples/QGRU_lime_class_2109} [are ...]\\
${\mathrm{limsse}^\mathrm{ms}_s}$&\input{examples/QGRU_lime_raw_2109} [are ...]\\
\gr${\mathrm{limsse}^\mathrm{ms}_p}$&\input{examples/QGRU_lime_prob_2109} [are ...]\\ \cline{1-2}
${\mathrm{grad}^\mathrm{L2}_{1 s}}$&\input{examples/LSTM_l2_raw_2109} [are ...]&\multirow{20}{*}{\rot{LSTM}}\\
\gr${\mathrm{grad}^\mathrm{L2}_{1 p}}$&\input{examples/LSTM_l2_prob_2109} [are ...]\\
${\mathrm{grad}^\mathrm{L2}_{\int s}}$&\input{examples/LSTM_l2_raw_integrated_2109} [are ...]\\
\gr${\mathrm{grad}^\mathrm{L2}_{\int p}}$&\input{examples/LSTM_l2_prob_integrated_2109} [are ...]\\
${\mathrm{grad}^\mathrm{dot}_{1 s}}$&\input{examples/LSTM_salience_raw_2109} [are ...]\\
\gr${\mathrm{grad}^\mathrm{dot}_{1 p}}$&\input{examples/LSTM_salience_prob_2109} [are ...] \\
${\mathrm{grad}^\mathrm{dot}_{\int s}}$&\input{examples/LSTM_salience_raw_integrated_2109} [are ...]\\
\gr${\mathrm{grad}^\mathrm{dot}_{\int p}}$&\input{examples/LSTM_salience_prob_integrated_2109} [are ...] \\
${\mathrm{omit}_1}$&\input{examples/LSTM_omission_2109} [are ...]\\
\gr${\mathrm{omit}_3}$&\input{examples/LSTM_omission-3_2109} [are ...]\\
${\mathrm{omit}_7}$&\input{examples/LSTM_omission-7_2109} [are ...]\\
\gr${\mathrm{occ}_1}$&\input{examples/LSTM_occlusion_2109} [are ...]\\
${\mathrm{occ}_3}$&\input{examples/LSTM_occlusion-3_2109} [are ...]\\
\gr${\mathrm{occ}_7}$&\input{examples/LSTM_occlusion-7_2109} [are ...]\\
${\mathrm{decomp}}$&\input{examples/LSTM_gamma_2109} [are ...]\\
\gr${\mathrm{lrp}}$&\input{examples/LSTM_lrp_2109} [are ...]\\
${\mathrm{deeplift}}$&\input{examples/LSTM_deeplift_2109} [are ...]\\
\gr${\mathrm{limsse}^\mathrm{bb}}$&\input{examples/LSTM_lime_class_2109} [are ...]\\
${\mathrm{limsse}^\mathrm{ms}_s}$&\input{examples/LSTM_lime_raw_2109} [are ...]\\
\gr${\mathrm{limsse}^\mathrm{ms}_p}$&\input{examples/LSTM_lime_prob_2109} [are ...]\\ \cline{1-2}
${\mathrm{grad}^\mathrm{L2}_{1 s}}$&\input{examples/QLSTM_l2_raw_2109} [are ...] & \multirow{20}{*}{\rot{QLSTM}}\\
\gr${\mathrm{grad}^\mathrm{L2}_{1 p}}$&\input{examples/QLSTM_l2_prob_2109} [are ...]\\
${\mathrm{grad}^\mathrm{L2}_{\int s}}$&\input{examples/QLSTM_l2_raw_integrated_2109} [are ...]\\
\gr${\mathrm{grad}^\mathrm{L2}_{\int p}}$&\input{examples/QLSTM_l2_prob_integrated_2109} [are ...]\\
${\mathrm{grad}^\mathrm{dot}_{1 s}}$&\input{examples/QLSTM_salience_raw_2109} [are ...]\\
\gr${\mathrm{grad}^\mathrm{dot}_{1 p}}$&\input{examples/QLSTM_salience_prob_2109} [are ...] \\
${\mathrm{grad}^\mathrm{dot}_{\int s}}$&\input{examples/QLSTM_salience_raw_integrated_2109} [are ...]\\
\gr${\mathrm{grad}^\mathrm{dot}_{\int p}}$&\input{examples/QLSTM_salience_prob_integrated_2109} [are ...] \\
${\mathrm{omit}_1}$&\input{examples/QLSTM_omission_2109} [are ...]\\
\gr${\mathrm{omit}_3}$&\input{examples/QLSTM_omission-3_2109} [are ...]\\
${\mathrm{omit}_7}$&\input{examples/QLSTM_omission-7_2109} [are ...]\\
\gr${\mathrm{occ}_1}$&\input{examples/QLSTM_occlusion_2109} [are ...]\\
${\mathrm{occ}_3}$&\input{examples/QLSTM_occlusion-3_2109} [are ...]\\
\gr${\mathrm{occ}_7}$&\input{examples/QLSTM_occlusion-7_2109} [are ...]\\
${\mathrm{decomp}}$&\input{examples/QLSTM_gamma_2109} [are ...]\\
\gr${\mathrm{lrp}}$&\input{examples/QLSTM_lrp_2109} [are ...]\\
${\mathrm{deeplift}}$&\input{examples/QLSTM_deeplift_2109} [are ...]\\
\gr${\mathrm{limsse}^\mathrm{bb}}$&\input{examples/QLSTM_lime_class_2109} [are ...]\\
${\mathrm{limsse}^\mathrm{ms}_s}$&\input{examples/QLSTM_lime_raw_2109} [are ...]\\
\gr${\mathrm{limsse}^\mathrm{ms}_p}$&\input{examples/QLSTM_lime_prob_2109} [are ...]
\end{tabular}
\caption{Verb context classified plural. Green (resp. red): evidence for (resp. against) the prediction.
Underlined: subject. Bold: $\mathrm{rmax}$ position.}
\label{fig:example-morphosyntactic1}
\end{figure}

%% file: example-morphosyntactic2.tex
\begin{figure}
\centering
\tiny
\begin{tabular}{lll}
${\mathrm{grad}^\mathrm{L2}_{1 s}}$&\input{examples/GRU_l2_raw_2446} [gives ...] & \multirow{20}{*}{\rot{GRU}}\\
\gr${\mathrm{grad}^\mathrm{L2}_{1 p}}$&\input{examples/GRU_l2_prob_2446} [gives ...]\\
${\mathrm{grad}^\mathrm{L2}_{\int s}}$&\input{examples/GRU_l2_raw_integrated_2446} [gives ...]\\
\gr${\mathrm{grad}^\mathrm{L2}_{\int p}}$&\input{examples/GRU_l2_prob_integrated_2446} [gives ...]\\
${\mathrm{grad}^\mathrm{dot}_{1 s}}$&\input{examples/GRU_salience_raw_2446} [gives ...]\\
\gr${\mathrm{grad}^\mathrm{dot}_{1 p}}$&\input{examples/GRU_salience_prob_2446} [gives ...] \\
${\mathrm{grad}^\mathrm{dot}_{\int s}}$&\input{examples/GRU_salience_raw_integrated_2446} [gives ...]\\
\gr${\mathrm{grad}^\mathrm{dot}_{\int p}}$&\input{examples/GRU_salience_prob_integrated_2446} [gives ...] \\
${\mathrm{omit}_1}$&\input{examples/GRU_omission_2446} [gives ...]\\
\gr${\mathrm{omit}_3}$&\input{examples/GRU_omission-3_2446} [gives ...]\\
${\mathrm{omit}_7}$&\input{examples/GRU_omission-7_2446} [gives ...]\\
\gr${\mathrm{occ}_1}$&\input{examples/GRU_occlusion_2446} [gives ...]\\
${\mathrm{occ}_3}$&\input{examples/GRU_occlusion-3_2446} [gives ...]\\
\gr${\mathrm{occ}_7}$&\input{examples/GRU_occlusion-7_2446} [gives ...]\\
${\mathrm{decomp}}$&\input{examples/GRU_gamma_2446} [gives ...]\\
\gr${\mathrm{lrp}}$&\input{examples/GRU_lrp_2446} [gives ...]\\
${\mathrm{deeplift}}$&\input{examples/GRU_deeplift_2446} [gives ...]\\
\gr${\mathrm{limsse}^\mathrm{bb}}$&\input{examples/GRU_lime_class_2446} [gives ...]\\
${\mathrm{limsse}^\mathrm{ms}_s}$&\input{examples/GRU_lime_raw_2446} [gives ...]\\
\gr${\mathrm{limsse}^\mathrm{ms}_p}$&\input{examples/GRU_lime_prob_2446} [gives ...]\\ \cline{1-2}
${\mathrm{grad}^\mathrm{L2}_{1 s}}$&\input{examples/QGRU_l2_raw_2446} [gives ...] & \multirow{20}{*}{\rot{QGRU}}\\
\gr${\mathrm{grad}^\mathrm{L2}_{1 p}}$&\input{examples/QGRU_l2_prob_2446} [gives ...]\\
${\mathrm{grad}^\mathrm{L2}_{\int s}}$&\input{examples/QGRU_l2_raw_integrated_2446} [gives ...]\\
\gr${\mathrm{grad}^\mathrm{L2}_{\int p}}$&\input{examples/QGRU_l2_prob_integrated_2446} [gives ...]\\
${\mathrm{grad}^\mathrm{dot}_{1 s}}$&\input{examples/QGRU_salience_raw_2446} [gives ...]\\
\gr${\mathrm{grad}^\mathrm{dot}_{1 p}}$&\input{examples/QGRU_salience_prob_2446} [gives ...] \\
${\mathrm{grad}^\mathrm{dot}_{\int s}}$&\input{examples/QGRU_salience_raw_integrated_2446} [gives ...]\\
\gr${\mathrm{grad}^\mathrm{dot}_{\int p}}$&\input{examples/QGRU_salience_prob_integrated_2446} [gives ...] \\
${\mathrm{omit}_1}$&\input{examples/QGRU_omission_2446} [gives ...]\\
\gr${\mathrm{omit}_3}$&\input{examples/QGRU_omission-3_2446} [gives ...]\\
${\mathrm{omit}_7}$&\input{examples/QGRU_omission-7_2446} [gives ...]\\
\gr${\mathrm{occ}_1}$&\input{examples/QGRU_occlusion_2446} [gives ...]\\
${\mathrm{occ}_3}$&\input{examples/QGRU_occlusion-3_2446} [gives ...]\\
\gr${\mathrm{occ}_7}$&\input{examples/QGRU_occlusion-7_2446} [gives ...]\\
${\mathrm{decomp}}$&\input{examples/QGRU_gamma_2446} [gives ...]\\
\gr${\mathrm{lrp}}$&\input{examples/QGRU_lrp_2446} [gives ...]\\
${\mathrm{deeplift}}$&\input{examples/QGRU_deeplift_2446} [gives ...]\\
\gr${\mathrm{limsse}^\mathrm{bb}}$&\input{examples/QGRU_lime_class_2446} [gives ...]\\
${\mathrm{limsse}^\mathrm{ms}_s}$&\input{examples/QGRU_lime_raw_2446} [gives ...]\\
\gr${\mathrm{limsse}^\mathrm{ms}_p}$&\input{examples/QGRU_lime_prob_2446} [gives ...] \\ \cline{1-2}
${\mathrm{grad}^\mathrm{L2}_{1 s}}$&\input{examples/LSTM_l2_raw_2446} [gives ...] & \multirow{20}{*}{\rot{LSTM}}\\
\gr${\mathrm{grad}^\mathrm{L2}_{1 p}}$&\input{examples/LSTM_l2_prob_2446} [gives ...]\\
${\mathrm{grad}^\mathrm{L2}_{\int s}}$&\input{examples/LSTM_l2_raw_integrated_2446} [gives ...]\\
\gr${\mathrm{grad}^\mathrm{L2}_{\int p}}$&\input{examples/LSTM_l2_prob_integrated_2446} [gives ...]\\
${\mathrm{grad}^\mathrm{dot}_{1 s}}$&\input{examples/LSTM_salience_raw_2446} [gives ...]\\
\gr${\mathrm{grad}^\mathrm{dot}_{1 p}}$&\input{examples/LSTM_salience_prob_2446} [gives ...] \\
${\mathrm{grad}^\mathrm{dot}_{\int s}}$&\input{examples/LSTM_salience_raw_integrated_2446} [gives ...]\\
\gr${\mathrm{grad}^\mathrm{dot}_{\int p}}$&\input{examples/LSTM_salience_prob_integrated_2446} [gives ...] \\
${\mathrm{omit}_1}$&\input{examples/LSTM_omission_2446} [gives ...]\\
\gr${\mathrm{omit}_3}$&\input{examples/LSTM_omission-3_2446} [gives ...]\\
${\mathrm{omit}_7}$&\input{examples/LSTM_omission-7_2446} [gives ...]\\
\gr${\mathrm{occ}_1}$&\input{examples/LSTM_occlusion_2446} [gives ...]\\
${\mathrm{occ}_3}$&\input{examples/LSTM_occlusion-3_2446} [gives ...]\\
\gr${\mathrm{occ}_7}$&\input{examples/LSTM_occlusion-7_2446} [gives ...]\\
${\mathrm{decomp}}$&\input{examples/LSTM_gamma_2446} [gives ...]\\
\gr${\mathrm{lrp}}$&\input{examples/LSTM_lrp_2446} [gives ...]\\
${\mathrm{deeplift}}$&\input{examples/LSTM_deeplift_2446} [gives ...]\\
\gr${\mathrm{limsse}^\mathrm{bb}}$&\input{examples/LSTM_lime_class_2446} [gives ...]\\
${\mathrm{limsse}^\mathrm{ms}_s}$&\input{examples/LSTM_lime_raw_2446} [gives ...]\\
\gr${\mathrm{limsse}^\mathrm{ms}_p}$&\input{examples/LSTM_lime_prob_2446} [gives ...]\\ \cline{1-2}
${\mathrm{grad}^\mathrm{L2}_{1 s}}$&\input{examples/QLSTM_l2_raw_2446} [gives ...] & \multirow{20}{*}{\rot{QLSTM}} \\
\gr${\mathrm{grad}^\mathrm{L2}_{1 p}}$&\input{examples/QLSTM_l2_prob_2446} [gives ...]\\
${\mathrm{grad}^\mathrm{L2}_{\int s}}$&\input{examples/QLSTM_l2_raw_integrated_2446} [gives ...]\\
\gr${\mathrm{grad}^\mathrm{L2}_{\int p}}$&\input{examples/QLSTM_l2_prob_integrated_2446} [gives ...]\\
${\mathrm{grad}^\mathrm{dot}_{1 s}}$&\input{examples/QLSTM_salience_raw_2446} [gives ...]\\
\gr${\mathrm{grad}^\mathrm{dot}_{1 p}}$&\input{examples/QLSTM_salience_prob_2446} [gives ...] \\
${\mathrm{grad}^\mathrm{dot}_{\int s}}$&\input{examples/QLSTM_salience_raw_integrated_2446} [gives ...]\\
\gr${\mathrm{grad}^\mathrm{dot}_{\int p}}$&\input{examples/QLSTM_salience_prob_integrated_2446} [gives ...] \\
${\mathrm{omit}_1}$&\input{examples/QLSTM_omission_2446} [gives ...]\\
\gr${\mathrm{omit}_3}$&\input{examples/QLSTM_omission-3_2446} [gives ...]\\
${\mathrm{omit}_7}$&\input{examples/QLSTM_omission-7_2446} [gives ...]\\
\gr${\mathrm{occ}_1}$&\input{examples/QLSTM_occlusion_2446} [gives ...]\\
${\mathrm{occ}_3}$&\input{examples/QLSTM_occlusion-3_2446} [gives ...]\\
\gr${\mathrm{occ}_7}$&\input{examples/QLSTM_occlusion-7_2446} [gives ...]\\
${\mathrm{decomp}}$&\input{examples/QLSTM_gamma_2446} [gives ...]\\
\gr${\mathrm{lrp}}$&\input{examples/QLSTM_lrp_2446} [gives ...]\\
${\mathrm{deeplift}}$&\input{examples/QLSTM_deeplift_2446} [gives ...]\\
\gr${\mathrm{limsse}^\mathrm{bb}}$&\input{examples/QLSTM_lime_class_2446} [gives ...]\\
${\mathrm{limsse}^\mathrm{ms}_s}$&\input{examples/QLSTM_lime_raw_2446} [gives ...]\\
\gr${\mathrm{limsse}^\mathrm{ms}_p}$&\input{examples/QLSTM_lime_prob_2446} [gives ...]
\end{tabular}
\caption{Verb context classified singular. Green (resp. red): evidence for (resp. against) the prediction.
Underlined: subject. Bold: $\mathrm{rmax}$ position.}
\label{fig:example-morphosyntactic2}
\end{figure}

%% file: example-morphosyntactic3.tex
\begin{figure*}
\centering
\scriptsize
\begin{tabular}{lll}
${\mathrm{grad}^\mathrm{L2}_{1 s}}$&\input{examples/GRU_l2_raw_1132} [has ...]&\multirow{20}{*}{\rot{GRU}}\\
\gr${\mathrm{grad}^\mathrm{L2}_{1 p}}$&\input{examples/GRU_l2_prob_1132} [has ...]\\
${\mathrm{grad}^\mathrm{L2}_{\int s}}$&\input{examples/GRU_l2_raw_integrated_1132} [has ...]\\
\gr${\mathrm{grad}^\mathrm{L2}_{\int p}}$&\input{examples/GRU_l2_prob_integrated_1132} [has ...]\\
${\mathrm{grad}^\mathrm{dot}_{1 s}}$&\input{examples/GRU_salience_raw_1132} [has ...]\\
\gr${\mathrm{grad}^\mathrm{dot}_{1 p}}$&\input{examples/GRU_salience_prob_1132} [has ...] \\
${\mathrm{grad}^\mathrm{dot}_{\int s}}$&\input{examples/GRU_salience_raw_integrated_1132} [has ...]\\
\gr${\mathrm{grad}^\mathrm{dot}_{\int p}}$&\input{examples/GRU_salience_prob_integrated_1132} [has ...] \\
${\mathrm{omit}_1}$&\input{examples/GRU_omission_1132} [has ...]\\
\gr${\mathrm{omit}_3}$&\input{examples/GRU_omission-3_1132} [has ...]\\
${\mathrm{omit}_7}$&\input{examples/GRU_omission-7_1132} [has ...]\\
\gr${\mathrm{occ}_1}$&\input{examples/GRU_occlusion_1132} [has ...]\\
${\mathrm{occ}_3}$&\input{examples/GRU_occlusion-3_1132} [has ...]\\
\gr${\mathrm{occ}_7}$&\input{examples/GRU_occlusion-7_1132} [has ...]\\
${\mathrm{decomp}}$&\input{examples/GRU_gamma_1132} [has ...]\\
\gr${\mathrm{lrp}}$&\input{examples/GRU_lrp_1132} [has ...]\\
${\mathrm{deeplift}}$&\input{examples/GRU_deeplift_1132} [has ...]\\
\gr${\mathrm{limsse}^\mathrm{bb}}$&\input{examples/GRU_lime_class_1132} [has ...]\\
${\mathrm{limsse}^\mathrm{ms}_s}$&\input{examples/GRU_lime_raw_1132} [has ...]\\
\gr${\mathrm{limsse}^\mathrm{ms}_p}$&\input{examples/GRU_lime_prob_1132} [has ...]\\ \cline{1-2}
${\mathrm{grad}^\mathrm{L2}_{1 s}}$&\input{examples/QLSTM_l2_raw_1132} [has ...]&\multirow{20}{*}{\rot{QLSTM}}\\
\gr${\mathrm{grad}^\mathrm{L2}_{1 p}}$&\input{examples/QLSTM_l2_prob_1132} [has ...]\\
${\mathrm{grad}^\mathrm{L2}_{\int s}}$&\input{examples/QLSTM_l2_raw_integrated_1132} [has ...]\\
\gr${\mathrm{grad}^\mathrm{L2}_{\int p}}$&\input{examples/QLSTM_l2_prob_integrated_1132} [has ...]\\
${\mathrm{grad}^\mathrm{dot}_{1 s}}$&\input{examples/QLSTM_salience_raw_1132} [has ...]\\
\gr${\mathrm{grad}^\mathrm{dot}_{1 p}}$&\input{examples/QLSTM_salience_prob_1132} [has ...] \\
${\mathrm{grad}^\mathrm{dot}_{\int s}}$&\input{examples/QLSTM_salience_raw_integrated_1132} [has ...]\\
\gr${\mathrm{grad}^\mathrm{dot}_{\int p}}$&\input{examples/QLSTM_salience_prob_integrated_1132} [has ...] \\
${\mathrm{omit}_1}$&\input{examples/QLSTM_omission_1132} [has ...]\\
\gr${\mathrm{omit}_3}$&\input{examples/QLSTM_omission-3_1132} [has ...]\\
${\mathrm{omit}_7}$&\input{examples/QLSTM_omission-7_1132} [has ...]\\
\gr${\mathrm{occ}_1}$&\input{examples/QLSTM_occlusion_1132} [has ...]\\
${\mathrm{occ}_3}$&\input{examples/QLSTM_occlusion-3_1132} [has ...]\\
\gr${\mathrm{occ}_7}$&\input{examples/QLSTM_occlusion-7_1132} [has ...]\\
${\mathrm{decomp}}$&\input{examples/QLSTM_gamma_1132} [has ...]\\
\gr${\mathrm{lrp}}$&\input{examples/QLSTM_lrp_1132} [has ...]\\
${\mathrm{deeplift}}$&\input{examples/QLSTM_deeplift_1132} [has ...]\\
\gr${\mathrm{limsse}^\mathrm{bb}}$&\input{examples/QLSTM_lime_class_1132} [has ...]\\
${\mathrm{limsse}^\mathrm{ms}_s}$&\input{examples/QLSTM_lime_raw_1132} [has ...]\\
\gr${\mathrm{limsse}^\mathrm{ms}_p}$&\input{examples/QLSTM_lime_prob_1132} [has ...] \\
\end{tabular}
\caption{Verb context classified singular by GRU and plural by QLSTM. Green (resp. red): evidence for (resp. against) the prediction. Underlined: subject. Bold: $\mathrm{rmax}$ position.}
\label{fig:example-morphosyntactic3}
\end{figure*}

%% file: example-manual-newsgroup-cnn.tex
\begin{figure*}
\center
\scriptsize
\begin{tabular}{m{0.025\textwidth}m{0.95\textwidth}}
\gr\rot{${{\mathrm{grad}^\mathrm{L2}_{1 s}}}$}& \input{examples/12_CNN_grad_raw_l2_138.tex} \\
\rot{${{\mathrm{grad}^\mathrm{L2}_{1 p}}}$}& \input{examples/12_CNN_grad_prob_l2_138.tex} \\
\gr\rot{${{\mathrm{grad}^\mathrm{L2}_{\int s}}}$}& \input{examples/12_CNN_grad_raw_l2_integrated_138.tex} \\
\rot{${{\mathrm{grad}^\mathrm{L2}_{\int p}}}$}& \input{examples/12_CNN_grad_prob_l2_integrated_138.tex} \\
\gr\rot{${{\mathrm{grad}^\mathrm{dot}_{1 s}}}$}& \input{examples/12_CNN_grad_raw_dot_138.tex} \\
\rot{${{\mathrm{grad}^\mathrm{dot}_{1 p}}}$}& \input{examples/12_CNN_grad_prob_dot_138.tex} \\
\gr\rot{${{\mathrm{grad}^\mathrm{dot}_{\int s}}}$}& \input{examples/12_CNN_grad_raw_dot_integrated_138.tex} \\
\rot{${{\mathrm{grad}^\mathrm{dot}_{\int p}}}$}& \input{examples/12_CNN_grad_prob_dot_integrated_138.tex} \\
\gr\rot{${{\mathrm{omit}_1}}$}& \input{examples/12_CNN_omission_138.tex} \\
\rot{${{\mathrm{omit}_3}}$}& \input{examples/12_CNN_omission-3_138.tex} \\
\gr\rot{${{\mathrm{omit}_7}}$}& \input{examples/12_CNN_omission-7_138.tex} \\
\rot{${{\mathrm{occ}_1}}$}& \input{examples/12_CNN_occlusion_138.tex} \\
\gr\rot{${{\mathrm{occ}_3}}$}& \input{examples/12_CNN_occlusion-3_138.tex} \\
\rot{${{\mathrm{occ}_7}}$}& \input{examples/12_CNN_occlusion-7_138.tex} \\
\gr\rot{${{\mathrm{lrp}}}$}& \input{examples/12_CNN_lrp_138.tex} \\
\rot{${{\mathrm{deeplift}}}$}& \input{examples/12_CNN_deeplift_138.tex} \\
\gr\rot{${{\mathrm{limsse}^\mathrm{bb}}}$}& \input{examples/12_CNN_lime_class_138.tex} \\
\rot{${{\mathrm{limsse}^\mathrm{ms}_s}}$}& \input{examples/12_CNN_lime_raw_138.tex} \\
\gr\rot{${{\mathrm{limsse}^\mathrm{ms}_p}}$}& \input{examples/12_CNN_lime_prob_138.tex} \\
\end{tabular}
\caption{{sci.electronics} post (not hybrid). Underlined: Manual relevance ground truth. Green (resp. red): evidence for (resp. against) {sci.electronics}.  Task method: CNN. Italics: OOV. Bold: $\mathrm{rmax}$ position.}
\label{fig:example-manual-newsgroup-lstm}
\end{figure*}

%% file: example-manual-newsgroup-gru.tex
\begin{figure*}
\center
\scriptsize
\begin{tabular}{m{0.025\textwidth}m{0.95\textwidth}}
\rot{${{\mathrm{grad}^\mathrm{L2}_{1 s}}}$}& \input{examples/13_GRU_grad_raw_l2_7.tex} \\
\gr\rot{${{\mathrm{grad}^\mathrm{L2}_{1 p}}}$}& \input{examples/13_GRU_grad_prob_l2_7.tex} \\
\rot{${{\mathrm{grad}^\mathrm{L2}_{\int s}}}$}& \input{examples/13_GRU_grad_raw_l2_integrated_7.tex} \\
\gr\rot{${{\mathrm{grad}^\mathrm{L2}_{\int p}}}$}& \input{examples/13_GRU_grad_prob_l2_integrated_7.tex} \\
\rot{${{\mathrm{grad}^\mathrm{dot}_{1 s}}}$}& \input{examples/13_GRU_grad_raw_dot_7.tex} \\
\gr\rot{${{\mathrm{grad}^\mathrm{dot}_{1 p}}}$}& \input{examples/13_GRU_grad_prob_dot_7.tex} \\
\rot{${{\mathrm{grad}^\mathrm{dot}_{\int s}}}$}& \input{examples/13_GRU_grad_raw_dot_integrated_7.tex} \\
\gr\rot{${{\mathrm{grad}^\mathrm{dot}_{\int p}}}$}& \input{examples/13_GRU_grad_prob_dot_integrated_7.tex} \\
\rot{${{\mathrm{omit}_1}}$}& \input{examples/13_GRU_omission_7.tex} \\
\gr\rot{${{\mathrm{omit}_3}}$}& \input{examples/13_GRU_omission-3_7.tex} \\
\rot{${{\mathrm{omit}_7}}$}& \input{examples/13_GRU_omission-7_7.tex} \\
\gr\rot{${{\mathrm{occ}_1}}$}& \input{examples/13_GRU_occlusion_7.tex} \\
\rot{${{\mathrm{occ}_3}}$}& \input{examples/13_GRU_occlusion-3_7.tex} \\
\gr\rot{${{\mathrm{occ}_7}}$}& \input{examples/13_GRU_occlusion-7_7.tex} \\
\rot{${{\mathrm{decomp}}}$}& \input{examples/13_GRU_gamma_7.tex} \\
\gr\rot{${{\mathrm{lrp}}}$}& \input{examples/13_GRU_lrp_7.tex} \\
\rot{${{\mathrm{deeplift}}}$}& \input{examples/13_GRU_deeplift_7.tex} \\
\gr\rot{${{\mathrm{limsse}^\mathrm{bb}}}$}& \input{examples/13_GRU_lime_class_7.tex} \\
\rot{${{\mathrm{limsse}^\mathrm{ms}_s}}$}& \input{examples/13_GRU_lime_raw_7.tex} \\
\gr\rot{${{\mathrm{limsse}^\mathrm{ms}_p}}$}& \input{examples/13_GRU_lime_prob_7.tex} \\
\end{tabular}
\caption{{sci.med} post (not hybrid). Underlined: Manual relevance ground truth. Green (resp. red): evidence for (resp. against) {sci.med}.  Task method: GRU. Italics: OOV. Bold: $\mathrm{rmax}$ position.}
\label{fig:example-manual-newsgroup-lstm}
\end{figure*}

%% file: example-hybrid-newsgroup-qgru.tex
\begin{figure*}
\center
\scriptsize
\begin{tabular}{m{0.025\textwidth}m{0.95\textwidth}}
\rot{${{\mathrm{grad}^\mathrm{L2}_{1 s}}}$}& \input{examples/QGRU_newsgroup_grad_raw_l2_630.tex} \\
\gr\rot{${{\mathrm{grad}^\mathrm{L2}_{1 p}}}$}& \input{examples/QGRU_newsgroup_grad_prob_l2_630.tex} \\
\rot{${{\mathrm{grad}^\mathrm{L2}_{\int s}}}$}& \input{examples/QGRU_newsgroup_grad_raw_l2_integrated_630.tex} \\
\gr\rot{${{\mathrm{grad}^\mathrm{L2}_{\int p}}}$}& \input{examples/QGRU_newsgroup_grad_prob_l2_integrated_630.tex} \\
\rot{${{\mathrm{grad}^\mathrm{dot}_{1 s}}}$}& \input{examples/QGRU_newsgroup_grad_raw_dot_630.tex} \\
\gr\rot{${{\mathrm{grad}^\mathrm{dot}_{1 p}}}$}& \input{examples/QGRU_newsgroup_grad_prob_dot_630.tex} \\
\rot{${{\mathrm{grad}^\mathrm{dot}_{\int s}}}$}& \input{examples/QGRU_newsgroup_grad_raw_dot_integrated_630.tex} \\
\gr\rot{${{\mathrm{grad}^\mathrm{dot}_{\int p}}}$}& \input{examples/QGRU_newsgroup_grad_prob_dot_integrated_630.tex} \\
\rot{${{\mathrm{omit}_1}}$}& \input{examples/QGRU_newsgroup_omission_630.tex} \\
\gr\rot{${{\mathrm{omit}_3}}$}& \input{examples/QGRU_newsgroup_omission-3_630.tex} \\
\rot{${{\mathrm{omit}_7}}$}& \input{examples/QGRU_newsgroup_omission-7_630.tex} \\
\gr\rot{${{\mathrm{occ}_1}}$}& \input{examples/QGRU_newsgroup_occlusion_630.tex} \\
\rot{${{\mathrm{occ}_3}}$}& \input{examples/QGRU_newsgroup_occlusion-3_630.tex} \\
\gr\rot{${{\mathrm{occ}_7}}$}& \input{examples/QGRU_newsgroup_occlusion-7_630.tex} \\
\rot{${{\mathrm{decomp}}}$}& \input{examples/QGRU_newsgroup_gamma_630.tex} \\
\gr\rot{${{\mathrm{lrp}}}$}& \input{examples/QGRU_newsgroup_lrp_630.tex} \\
\rot{${{\mathrm{deeplift}}}$}& \input{examples/QGRU_newsgroup_deeplift_630.tex} \\
\gr\rot{${{\mathrm{limsse}^\mathrm{bb}}}$}& \input{examples/QGRU_newsgroup_lime_class_630.tex} \\
\rot{${{\mathrm{limsse}^\mathrm{ms}_s}}$}& \input{examples/QGRU_newsgroup_lime_raw_630.tex} \\
\gr\rot{${{\mathrm{limsse}^\mathrm{ms}_p}}$}& \input{examples/QGRU_newsgroup_lime_prob_630.tex} \\
\end{tabular}
\caption{Hybrid newsgroup post, classified {talk.politics.mideast}. Green (resp. red): evidence for (resp. against) {talk.politics.mideast}. Underlined: {talk.politics.mideast} fragment. Italics: OOV. Task method: QGRU. Bold: $\mathrm{rmax}$ position.}
\label{fig:example-hybrid-newsgroup-QGRU}
\end{figure*}

%% file: example-hybrid-newsgroup-lstm.tex
\begin{figure*}
\center
\scriptsize
\begin{tabular}{m{0.025\textwidth}m{0.95\textwidth}}
\rot{${{\mathrm{grad}^\mathrm{L2}_{1 s}}}$}& \input{examples/LSTM_newsgroup_grad_raw_l2_1028.tex} \\
\gr\rot{${{\mathrm{grad}^\mathrm{L2}_{1 p}}}$}& \input{examples/LSTM_newsgroup_grad_prob_l2_1028.tex} \\
\rot{${{\mathrm{grad}^\mathrm{L2}_{\int s}}}$}& \input{examples/LSTM_newsgroup_grad_raw_l2_integrated_1028.tex} \\
\gr\rot{${{\mathrm{grad}^\mathrm{L2}_{\int p}}}$}& \input{examples/LSTM_newsgroup_grad_prob_l2_integrated_1028.tex} \\
\rot{${{\mathrm{grad}^\mathrm{dot}_{1 s}}}$}& \input{examples/LSTM_newsgroup_grad_raw_dot_1028.tex} \\
\gr\rot{${{\mathrm{grad}^\mathrm{dot}_{1 p}}}$}& \input{examples/LSTM_newsgroup_grad_prob_dot_1028.tex} \\
\rot{${{\mathrm{grad}^\mathrm{dot}_{\int s}}}$}& \input{examples/LSTM_newsgroup_grad_raw_dot_integrated_1028.tex} \\
\gr\rot{${{\mathrm{grad}^\mathrm{dot}_{\int p}}}$}& \input{examples/LSTM_newsgroup_grad_prob_dot_integrated_1028.tex} \\
\rot{${{\mathrm{omit}_1}}$}& \input{examples/LSTM_newsgroup_omission_1028.tex} \\
\gr\rot{${{\mathrm{omit}_3}}$}& \input{examples/LSTM_newsgroup_omission-3_1028.tex} \\
\rot{${{\mathrm{omit}_7}}$}& \input{examples/LSTM_newsgroup_omission-7_1028.tex} \\
\gr\rot{${{\mathrm{occ}_1}}$}& \input{examples/LSTM_newsgroup_occlusion_1028.tex} \\
\rot{${{\mathrm{occ}_3}}$}& \input{examples/LSTM_newsgroup_occlusion-3_1028.tex} \\
\gr\rot{${{\mathrm{occ}_7}}$}& \input{examples/LSTM_newsgroup_occlusion-7_1028.tex} \\
\rot{${{\mathrm{decomp}}}$}& \input{examples/LSTM_newsgroup_gamma_1028.tex} \\
\gr\rot{${{\mathrm{lrp}}}$}& \input{examples/LSTM_newsgroup_lrp_1028.tex} \\
\rot{${{\mathrm{deeplift}}}$}& \input{examples/LSTM_newsgroup_deeplift_1028.tex} \\
\gr\rot{${{\mathrm{limsse}^\mathrm{bb}}}$}& \input{examples/LSTM_newsgroup_lime_class_1028.tex} \\
\rot{${{\mathrm{limsse}^\mathrm{ms}_s}}$}& \input{examples/LSTM_newsgroup_lime_raw_1028.tex} \\
\gr\rot{${{\mathrm{limsse}^\mathrm{ms}_p}}$}& \input{examples/LSTM_newsgroup_lime_prob_1028.tex} \\
\end{tabular}
\caption{Hybrid newsgroup post, classified comp.windows.x. Green (resp. red): evidence for (resp. against) comp.windows.x. Underlined: comp.windows.x fragment. Italics: OOV. Task method: LSTM. Bold: $\mathrm{rmax}$ position. The telephone numbers in the last sentence appear in 3 comp.windows.x posts but nowhere else in the corpus.}
\label{fig:example-hybrid-newsgroup-LSTM}
\end{figure*}

%% file: example-hybrid-newsgroup-qlstm.tex
\begin{figure*}
\center
\scriptsize
\begin{tabular}{m{0.025\textwidth}m{0.95\textwidth}}
\rot{${{\mathrm{grad}^\mathrm{L2}_{1 s}}}$}& \input{examples/QLSTM_newsgroup_grad_raw_l2_3313.tex} \\
\gr\rot{${{\mathrm{grad}^\mathrm{L2}_{1 p}}}$}& \input{examples/QLSTM_newsgroup_grad_prob_l2_3313.tex} \\
\rot{${{\mathrm{grad}^\mathrm{L2}_{\int s}}}$}& \input{examples/QLSTM_newsgroup_grad_raw_l2_integrated_3313.tex} \\
\gr\rot{${{\mathrm{grad}^\mathrm{L2}_{\int p}}}$}& \input{examples/QLSTM_newsgroup_grad_prob_l2_integrated_3313.tex} \\
\rot{${{\mathrm{grad}^\mathrm{dot}_{1 s}}}$}& \input{examples/QLSTM_newsgroup_grad_raw_dot_3313.tex} \\
\gr\rot{${{\mathrm{grad}^\mathrm{dot}_{1 p}}}$}& \input{examples/QLSTM_newsgroup_grad_prob_dot_3313.tex} \\
\rot{${{\mathrm{grad}^\mathrm{dot}_{\int s}}}$}& \input{examples/QLSTM_newsgroup_grad_raw_dot_integrated_3313.tex} \\
\gr\rot{${{\mathrm{grad}^\mathrm{dot}_{\int p}}}$}& \input{examples/QLSTM_newsgroup_grad_prob_dot_integrated_3313.tex} \\
\rot{${{\mathrm{omit}_1}}$}& \input{examples/QLSTM_newsgroup_omission_3313.tex} \\
\gr\rot{${{\mathrm{omit}_3}}$}& \input{examples/QLSTM_newsgroup_omission-3_3313.tex} \\
\rot{${{\mathrm{omit}_7}}$}& \input{examples/QLSTM_newsgroup_omission-7_3313.tex} \\
\gr\rot{${{\mathrm{occ}_1}}$}& \input{examples/QLSTM_newsgroup_occlusion_3313.tex} \\
\rot{${{\mathrm{occ}_3}}$}& \input{examples/QLSTM_newsgroup_occlusion-3_3313.tex} \\
\gr\rot{${{\mathrm{occ}_7}}$}& \input{examples/QLSTM_newsgroup_occlusion-7_3313.tex} \\
\rot{${{\mathrm{decomp}}}$}& \input{examples/QLSTM_newsgroup_gamma_3313.tex} \\
\gr\rot{${{\mathrm{lrp}}}$}& \input{examples/QLSTM_newsgroup_lrp_3313.tex} \\
\rot{${{\mathrm{deeplift}}}$}& \input{examples/QLSTM_newsgroup_deeplift_3313.tex} \\
\gr\rot{${{\mathrm{limsse}^\mathrm{bb}}}$}& \input{examples/QLSTM_newsgroup_lime_class_3313.tex} \\
\rot{${{\mathrm{limsse}^\mathrm{ms}_s}}$}& \input{examples/QLSTM_newsgroup_lime_raw_3313.tex} \\
\gr\rot{${{\mathrm{limsse}^\mathrm{ms}_p}}$}& \input{examples/QLSTM_newsgroup_lime_prob_3313.tex} \\
\end{tabular}
\caption{Hybrid newsgroup post, classified comp.windows.x. Green (resp. red): evidence for (resp. against) comp.windows.x. Underlined: comp.windows.x fragment. Italics: OOV. Task method: QLSTM. Bold: $\mathrm{rmax}$ position.}
\label{fig:example-hybrid-newsgroup-GRU}
\end{figure*}

%% file: example-hybrid-yelp-gru.tex
\begin{figure*}
\center
\scriptsize
\begin{tabular}{m{0.025\textwidth}m{0.95\textwidth}}
\rot{${{\mathrm{grad}^\mathrm{L2}_{1 s}}}$}& \input{examples/GRU_yelp_grad_raw_l2_3707.tex} \\
\gr\rot{${{\mathrm{grad}^\mathrm{L2}_{1 p}}}$}& \input{examples/GRU_yelp_grad_prob_l2_3707.tex} \\
\rot{${{\mathrm{grad}^\mathrm{L2}_{\int s}}}$}& \input{examples/GRU_yelp_grad_raw_l2_integrated_3707.tex} \\
\gr\rot{${{\mathrm{grad}^\mathrm{L2}_{\int p}}}$}& \input{examples/GRU_yelp_grad_prob_l2_integrated_3707.tex} \\
\rot{${{\mathrm{grad}^\mathrm{dot}_{1 s}}}$}& \input{examples/GRU_yelp_grad_raw_dot_3707.tex} \\
\gr\rot{${{\mathrm{grad}^\mathrm{dot}_{1 p}}}$}& \input{examples/GRU_yelp_grad_prob_dot_3707.tex} \\
\rot{${{\mathrm{grad}^\mathrm{dot}_{\int s}}}$}& \input{examples/GRU_yelp_grad_raw_dot_integrated_3707.tex} \\
\gr\rot{${{\mathrm{grad}^\mathrm{dot}_{\int p}}}$}& \input{examples/GRU_yelp_grad_prob_dot_integrated_3707.tex} \\
\rot{${{\mathrm{omit}_1}}$}& \input{examples/GRU_yelp_omission_3707.tex} \\
\gr\rot{${{\mathrm{omit}_3}}$}& \input{examples/GRU_yelp_omission-3_3707.tex} \\
\rot{${{\mathrm{omit}_7}}$}& \input{examples/GRU_yelp_omission-7_3707.tex} \\
\gr\rot{${{\mathrm{occ}_1}}$}& \input{examples/GRU_yelp_occlusion_3707.tex} \\
\rot{${{\mathrm{occ}_3}}$}& \input{examples/GRU_yelp_occlusion-3_3707.tex} \\
\gr\rot{${{\mathrm{occ}_7}}$}& \input{examples/GRU_yelp_occlusion-7_3707.tex} \\
\rot{${{\mathrm{decomp}}}$}& \input{examples/GRU_yelp_gamma_3707.tex} \\
\gr\rot{${{\mathrm{lrp}}}$}& \input{examples/GRU_yelp_lrp_3707.tex} \\
\rot{${{\mathrm{deeplift}}}$}& \input{examples/GRU_yelp_deeplift_3707.tex} \\
\gr\rot{${{\mathrm{limsse}^\mathrm{bb}}}$}& \input{examples/GRU_yelp_lime_class_3707.tex} \\
\rot{${{\mathrm{limsse}^\mathrm{ms}_s}}$}& \input{examples/GRU_yelp_lime_raw_3707.tex} \\
\gr\rot{${{\mathrm{limsse}^\mathrm{ms}_p}}$}& \input{examples/GRU_yelp_lime_prob_3707.tex} \\
\end{tabular}
\caption{Hybrid yelp review, classified {positive}. Green (resp. red): evidence for (resp. against) {positive}. Underlined: {positive} fragments. Italics: OOV. Task method: GRU. Bold: $\mathrm{rmax}$ position.}
\label{fig:example-hybrid-yelp-gru}
\end{figure*}

%% file: example-hybrid-yelp-lstm.tex
\begin{figure*}
\center
\scriptsize
\begin{tabular}{m{0.025\textwidth}m{0.95\textwidth}}
\rot{${{\mathrm{grad}^\mathrm{L2}_{1 s}}}$}& \input{examples/LSTM_yelp_grad_raw_l2_5667.tex} \\
\gr\rot{${{\mathrm{grad}^\mathrm{L2}_{1 p}}}$}& \input{examples/LSTM_yelp_grad_prob_l2_5667.tex} \\
\rot{${{\mathrm{grad}^\mathrm{L2}_{\int s}}}$}& \input{examples/LSTM_yelp_grad_raw_l2_integrated_5667.tex} \\
\gr\rot{${{\mathrm{grad}^\mathrm{L2}_{\int p}}}$}& \input{examples/LSTM_yelp_grad_prob_l2_integrated_5667.tex} \\
\rot{${{\mathrm{grad}^\mathrm{dot}_{1 s}}}$}& \input{examples/LSTM_yelp_grad_raw_dot_5667.tex} \\
\gr\rot{${{\mathrm{grad}^\mathrm{dot}_{1 p}}}$}& \input{examples/LSTM_yelp_grad_prob_dot_5667.tex} \\
\rot{${{\mathrm{grad}^\mathrm{dot}_{\int s}}}$}& \input{examples/LSTM_yelp_grad_raw_dot_integrated_5667.tex} \\
\gr\rot{${{\mathrm{grad}^\mathrm{dot}_{\int p}}}$}& \input{examples/LSTM_yelp_grad_prob_dot_integrated_5667.tex} \\
\rot{${{\mathrm{omit}_1}}$}& \input{examples/LSTM_yelp_omission_5667.tex} \\
\gr\rot{${{\mathrm{omit}_3}}$}& \input{examples/LSTM_yelp_omission-3_5667.tex} \\
\rot{${{\mathrm{omit}_7}}$}& \input{examples/LSTM_yelp_omission-7_5667.tex} \\
\gr\rot{${{\mathrm{occ}_1}}$}& \input{examples/LSTM_yelp_occlusion_5667.tex} \\
\rot{${{\mathrm{occ}_3}}$}& \input{examples/LSTM_yelp_occlusion-3_5667.tex} \\
\gr\rot{${{\mathrm{occ}_7}}$}& \input{examples/LSTM_yelp_occlusion-7_5667.tex} \\
\rot{${{\mathrm{decomp}}}$}& \input{examples/LSTM_yelp_gamma_5667.tex} \\
\gr\rot{${{\mathrm{lrp}}}$}& \input{examples/LSTM_yelp_lrp_5667.tex} \\
\rot{${{\mathrm{deeplift}}}$}& \input{examples/LSTM_yelp_deeplift_5667.tex} \\
\gr\rot{${{\mathrm{limsse}^\mathrm{bb}}}$}& \input{examples/LSTM_yelp_lime_class_5667.tex} \\
\rot{${{\mathrm{limsse}^\mathrm{ms}_s}}$}& \input{examples/LSTM_yelp_lime_raw_5667.tex} \\
\gr\rot{${{\mathrm{limsse}^\mathrm{ms}_p}}$}& \input{examples/LSTM_yelp_lime_prob_5667.tex} \\
\end{tabular}
\caption{Hybrid yelp review, classified {negative}. Green (resp. red): evidence for (resp. against) {negative}. Underlined: {negative} fragments. Italics: OOV. Task method: LSTM. Bold: $\mathrm{rmax}$ position.}
\label{fig:example-hybrid-yelp-LSTM}
\end{figure*}